\documentclass[acmsmall]{acmart}

\AtBeginDocument{%
  \providecommand\BibTeX{{%
    \normalfont B\kern-0.5em{\scshape i\kern-0.25em b}\kern-0.8em\TeX}}}

\setcopyright{acmcopyright}
\copyrightyear{2023}
\acmYear{2023}
\acmDOI{FIXME}

\acmJournal{TALLIP}
\acmVolume{000}
\acmNumber{000}
\acmArticle{000}
\acmMonth{000}

\setcopyright{acmlicensed}
\acmJournal{TALLIP}
\acmYear{2023} \acmVolume{1} \acmNumber{1} \acmArticle{1} \acmMonth{1} \acmPrice{15.00}\acmDOI{10.1145/3610611}

\usepackage{subfigure}
\usepackage{graphicx}

\usepackage{times}
\usepackage{soul}
\usepackage[utf8]{inputenc}
\usepackage{bm}
\usepackage[linesnumbered,ruled,vlined]{algorithm2e}

\usepackage{amsmath}
\usepackage{amsthm}
\usepackage{mathtools}
\usepackage{algpseudocode}

\newcommand{\zstroke}{%
  \text{\ooalign{\hidewidth -\kern-.3em-\hidewidth\cr$z$\cr}}%
}

\usepackage{url}
\usepackage{hyperref}
\hypersetup
{
    colorlinks=true,
    citecolor=navyblue,
    linkcolor=navyblue,
    filecolor=navyblue,
    urlcolor=navyblue,
}

\usepackage{tabularx}
\usepackage{booktabs}
\usepackage{makecell}
\usepackage{diagbox}
\usepackage{multirow}
\usepackage{caption}
\usepackage[detect-all]{siunitx}

\usepackage{etoolbox} 
\robustify\bfseries
\newcommand{\superscript}[1]{\ensuremath{^{\textrm{#1}}}}

\usepackage{natbib}

\newcommand{\fn}[1]{\footnote{#1}}
\usepackage{lipsum}

\usepackage{CJKutf8}

\usepackage{comment}

\usepackage{verbatim}

\usepackage{color}
\usepackage{colortbl}
\usepackage{xcolor}
\usepackage{tabu}
\definecolor{cR}{RGB}{255,204,204}
\definecolor{cG}{RGB}{204,255,204}
\definecolor{cB}{RGB}{204,204,255}
\definecolor{cK}{RGB}{221,221,221}
\definecolor{navyblue}{rgb}{0.0,0.0,0.5}
\definecolor{lB}{rgb}{0,0.4,0.9}

\newcommand{\Sec}[1]{Section~\ref{sec:#1}}
\newcommand{\Fig}[1]{Figure~\ref{fig:#1}}
\newcommand{\Tab}[1]{Table~\ref{tab:#1}}
\newcommand{\Algo}[1]{Algorithm~\ref{algo:#1}}
\newcommand{\Eq}[1]{Eq.~(\ref{eq:#1})}
\newcommand{\ra}{{$\rightarrow$}}

\usepackage{soul}
\sethlcolor{blue}
\renewcommand{\hl}[2][blue]{\textcolor{#1}{#2}}

\renewcommand{\hl}[1]{#1}
\definecolor{blue}{rgb}{0,0,0}

\makeatletter
\def\adl@drawiv#1#2#3{%
        \hskip.5\tabcolsep
        \xleaders#3{#2.5\@tempdimb #1{1}#2.5\@tempdimb}%
                #2\z@ plus1fil minus1fil\relax
        \hskip.5\tabcolsep}
\newcommand{\cdashlinelr}[1]{%
  \noalign{\vskip\aboverulesep
           \global\let\@dashdrawstore\adl@draw
           \global\let\adl@draw\adl@drawiv}
  \cdashline{#1}
  \noalign{\global\let\adl@draw\@dashdrawstore
           \vskip\belowrulesep}}
\makeatother

\usepackage{microtype}

\begin{document}
\title[SelfSeg: A Self-supervised Sub-word Segmentation Method for Neural Machine Translation]{SelfSeg: A Self-supervised Sub-word Segmentation Method for Neural Machine Translation}

\author{Haiyue Song}
\email{song@nlp.ist.i.kyoto-u.ac.jp}
\affiliation
{
    \institution{Kyoto University}
    \country{Japan}
}
\author{Raj Dabre}
\email{raj.dabre@nict.go.jp}
\affiliation
{
    \institution{National Institute of Information and Communications Technology}
    \country{Japan}
}
\author{Chenhui Chu}
\email{chu@i.kyoto-u.ac.jp}
\affiliation
{
    \institution{Kyoto University}
    \country{Japan}
}

\author{Sadao Kurohashi}
\email{kuro@i.kyoto-u.ac.jp}
\affiliation
{
    \institution{Kyoto University}
    \country{Japan}
}
\author{Eiichiro Sumita}
\email{eiichiro.sumita@nict.go.jp}
\affiliation
{
    \institution{National Institute of Information and Communications Technology}
    \country{Japan}
}
\renewcommand{\shortauthors}{Song, et al.}


\begin{abstract}
Sub-word segmentation is an essential pre-processing step for Neural Machine Translation (NMT). Existing work has shown that neural sub-word segmenters are better than Byte-Pair Encoding (BPE), however, they are inefficient as they require parallel corpora, days to train and hours to decode. This paper introduces SelfSeg, a self-supervised neural sub-word segmentation method that is much faster to train/decode and requires only monolingual dictionaries instead of parallel corpora. SelfSeg takes as input a word in the form of a partially masked character sequence, optimizes the word generation probability and generates the segmentation with the maximum posterior probability, which is calculated using a dynamic programming algorithm. The training time of SelfSeg depends on word frequencies, and we explore several word frequency normalization strategies to accelerate the training phase. Additionally, we propose a regularization mechanism that allows the segmenter to generate various segmentations for one word. To show the effectiveness of our approach, we conduct MT experiments in low-, middle- and high-resource scenarios, where we compare the performance of using different segmentation methods. The experimental results demonstrate that on the low-resource ALT dataset, our method achieves more than $1.2$ BLEU score improvement compared with BPE and SentencePiece, and a $1.1$ score improvement over Dynamic Programming Encoding (DPE) and Vocabulary Learning via Optimal Transport (VOLT) on average. The regularization method achieves approximately a $4.3$ BLEU score improvement over BPE and a $1.2$ BLEU score improvement over BPE-dropout, the regularized version of BPE. We also observed significant improvements on IWSLT15 Vi\ra En, WMT16 Ro\ra En and WMT15 Fi\ra En datasets, and competitive results on the WMT14 De\ra En \hl{and WMT14 Fr\ra En} datasets. Furthermore, our method is $17.8$x faster during training and up to $36.8$x faster during decoding in a high-resource scenario compared to DPE. We provide extensive analysis, including why monolingual word-level data is enough to train SelfSeg.
\end{abstract}

\maketitle
\section{Introduction}
\label{sec:introduction}

NMT is the most prevalent approach for machine translation~\cite{NIPS2014_5346,2014arXiv1409.0473B,NIPS2017_3f5ee243,10.5555/3305381.3305510} due to its end-to-end nature and its ability to achieve state-of-the-art translations. 
Early NMT methods consider words as the minimal input unit and use a vocabulary to hold frequent words~~\cite{NIPS2014_5346,2014arXiv1409.0473B,luong-etal-2015-addressing, kalchbrenner-blunsom-2013-recurrent-continuous}. However, they face the out-of-vocabulary (OOV) problem due to the limited size of the vocabulary and the unlimited variety of words in the test data. Even with a very large vocabulary that covers most words in the train set, for morphologically rich languages such as German, there are still $3\%$ of new types of words that appear in the test set ~\cite{jean-etal-2015-using}. This largely hinders the translation quality of sentences with many rare words~\cite{NIPS2014_5346, 2014arXiv1409.0473B}.

Sub-word segmentation is dedicated to addressing the OOV problem by segmenting rare words into sub-words or characters that are present in a vocabulary. Frequency-based methods first use a monolingual corpus to build a sub-word vocabulary that contains characters, high-frequency sub-word fragments and common words. During decoding, for each word or sentence, it recursively combines an adjacent fragment pair that occurs most frequently according to the sub-word vocabulary, starting from characters~\citep{sennrich-etal-2016-neural, kudo-richardson-2018-sentencepiece}. The main limitation is that these segmentation methods are not optimized for downstream tasks, such as NMT.
DPE~\citep{he-etal-2020-dynamic}, a recently proposed neural sub-word segmentation approach, views the target sentence as a latent variable whose probability is the sum of the probability of all possible segmentations. The probability of each segmentation is calculated by a transformer model conditioned on the source sentence. It optimizes the target sentence probability in the training phase and outputs the segmentation with maximum posterior probability in the decoding phase. The DPE work also shows the importance of optimizing sub-word segmentation for the MT task.
Different from BPE~\cite{sennrich-etal-2016-neural}, it uses parallel data and deploys a neural sequence-to-sequence model for the segmentation. This is a double-edged sword: on one hand, using a sequence-to-sequence neural model enables the segmentation to be aware of all past tokens where BPE does not. Because the NMT decoder is also aware of all past tokens, this segmentation approach may be optimal; on the other hand, it is not practical neither in low-resource scenarios where large parallel corpora are not available, nor in high-resource scenarios where training and decoding take hours to days. 

\begin{figure}
\centering  
\subfigure[Japanese word segmentation task. The sentence is consistently segmented with different document-level contexts. For the context we show the translated English references only.]{\label{fig:word_segmentation_example}\includegraphics[width=0.9\columnwidth]{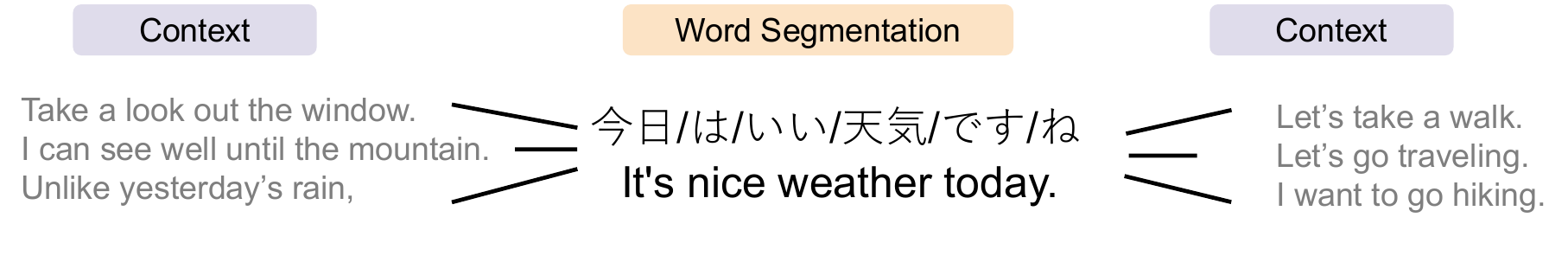}}
\subfigure[English sub-word segmentation task. The word is consistently segmented with different sentence-level contexts.]{\label{fig:sub-word_segmentation_example}\includegraphics[width=0.85\columnwidth]{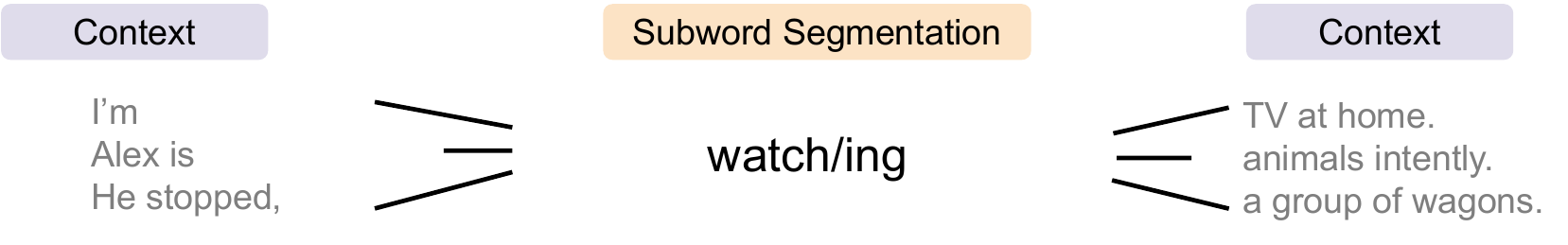}}
\caption{Segmentation is a self-contained task where context information is not required.}
\end{figure}


Leveraging existing large-scale monolingual data through self-supervised learning methods significantly reduces the need for parallel corpora. Predicting masked tokens is a promising task to provide training signals for an encoder that could be fine-tuned for a variety of downstream tasks~\cite{devlin-etal-2019-bert}, or an encoder-decoder model which could boost the MT tasks \cite{2019arXiv190502450S}. Although relying on monolingual data obviates the need for parallel corpora, the DPE method will still be slow as entire sequences have to be processed. In order to speed up the model, we propose that words be used instead of sentences. The motivation comes from the examples in Figures~\ref{fig:word_segmentation_example} and ~\ref{fig:sub-word_segmentation_example}.
In~\Fig{word_segmentation_example}, for a Japanese word segmentation task, the sentence will be consistently segmented in different document-level contexts. It is similar for sub-word segmentation as presented in~\Fig{sub-word_segmentation_example}, where we don't need sentence-level information.
For example, the word ``watching'' should be consistently segmented into ``watch+ing'' no matter which sentence the word is in.
This insight can help us go from sentence-level data to word-level data to train the sub-word segmenter, which significantly improves the training and decoding speed, because the training requires only word-level data and one type of word needs to be decoded only once. 

Based on these observations, we propose SelfSeg, a sub-word segmenter that trained on monolingual word-level data.
It uses a neural model to optimize the word generation probability conditioned on partially masked words, and outputs the segmentation with the maximum posterior probability. The decoding is fast because it only needs to decode each unique word once. To speed up the training phase, we propose a word frequency normalization method that adjusts the frequencies for frequent and rare words. Furthermore, motivated by \citet{provilkov-etal-2020-bpe} we also implement a regularization method on top of SelfSeg which provides multiple segmentations of the same word. We conduct experiments for low-, middle- and high-resource language pairs using the corpora from Asian Language Treebank (ALT), IWSLT and WMT. We show that SelfSeg yields segmentations that achieve better translation quality of up to 1.1-1.3 BLEU compared to existing approaches such as BPE~\cite{sennrich-etal-2016-neural}, SentencePiece~\cite{kudo-2018-subword}, DPE~\citep{he-etal-2020-dynamic} and VOLT~\cite{xu-etal-2021-vocabulary}. Additionally, we show that in low-resource settings regularized SelfSeg not only outperforms BPE by 4.3 BLEU but also BPE-dropout~\cite{provilkov-etal-2020-bpe} by 1.2 BLEU. We also provide analyses exploring various aspects of SelfSeg. 
Our contributions are as follows:
\begin{itemize}
    \item \textbf{We propose SelfSeg}, a neural sub-word segmentation method that relies on only monolingual word-level data with masking strategies, together with word-frequency normalization strategies to speed up the training, and a regularization mechanism.
    \item \textbf{Experimental results show significant BLEU score improvements} over existing works, as well as a significant increase in training and decoding speed compared to neural approaches such as DPE.
    \item \textbf{We provide extensive analysis}, including the effect of different masking methods and normalization methods, and why monolingual word-level data is enough to train SelfSeg.
\end{itemize}
\begin{figure}
\centering  
\subfigure[During the training phase model maximizes the probability of one word.]{\label{fig:sub-word_segmentation_train}\includegraphics[width=0.45\columnwidth]{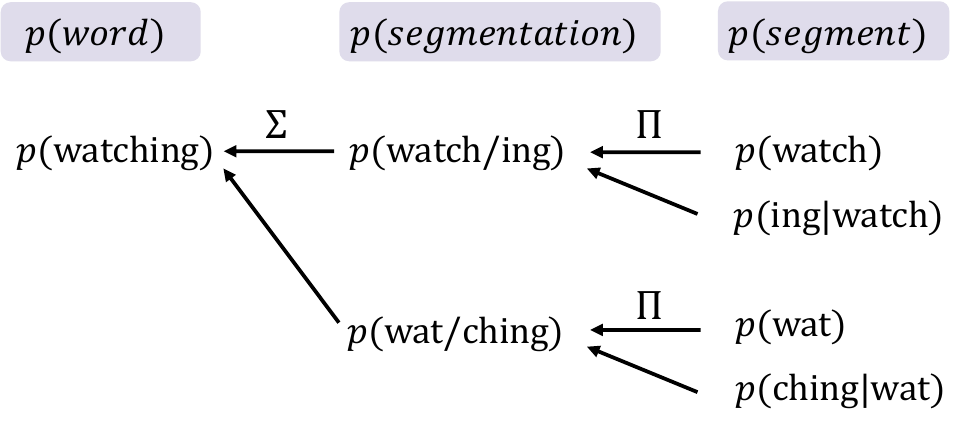}}
\hspace{7pt}
\subfigure[During the decoding phase model seeks the segmentation with the highest probability.]{\label{fig:sub-word_segmentation_test}\includegraphics[width=0.45\columnwidth]{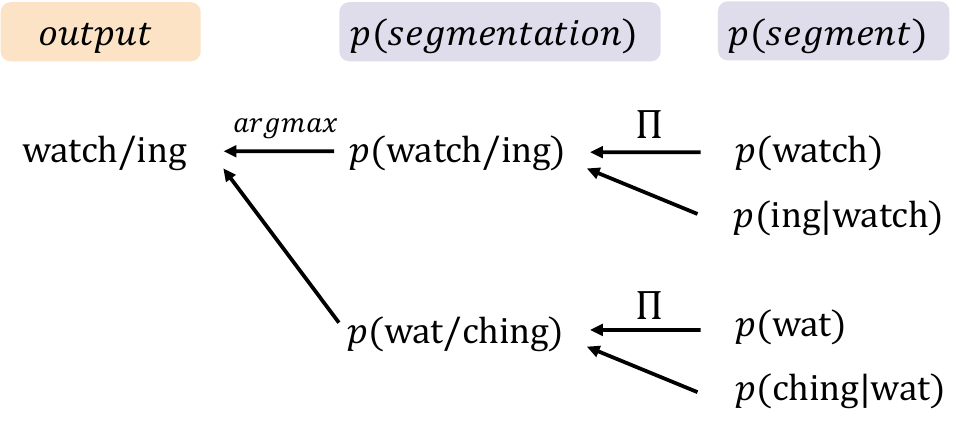}}
\caption{The training and decoding steps for sub-word segmentation.}
\label{fig:sub-word_segmentation_train_test}
\end{figure}
\section{Related Work}
In this section, we introduce two categories of sub-word segmentation methods, namely, non-neural and neural methods. In addition, we introduce the prevalent self-supervised learning paradigm.
\subsection{Non-Neural Sub-word Segmentation}
Initial works on NMT used word-level vocabularies that could only represent the most frequent words, leading to the OOV problem \cite{NIPS2014_5346}. Character-based or byte-based approaches solve the OOV problem, however, they introduce higher computational complexity and thus translation latency because they generate longer sequences and require deeper-stacked models (often equipped with pre-layer normalization)~\cite{1911.04997, Kim_Jernite_Sontag_Rush_2016, costa-jussa-fonollosa-2016-character, 1511.04586, 1604.00788, cherry-etal-2018-revisiting,shaham-levy-2021-neural}. Fully character-based NMT systems show higher translation quality compared with word-based systems, especially for morphologically rich languages~\cite{Kim_Jernite_Sontag_Rush_2016, costa-jussa-fonollosa-2016-character, 1511.04586}, while a hybrid word-character model shows a larger improvement~\cite{1604.00788}. A recent study further represents every computerized text as a sequence of bytes via UTF-8~\cite{shaham-levy-2021-neural}. 

Sub-word segmentation methods address both the OOV problem and the computational cost of the character-based methods, thus becoming an indispensable pre-processing step for modern NMT models~\cite{sennrich-etal-2016-neural, provilkov-etal-2020-bpe, 1909.03341, 6289079, kudo-richardson-2018-sentencepiece, kudo-2018-subword}. \citet{sennrich-etal-2016-neural} adapt BPE compression algorithm ~\cite{gage1994new} to the task of sub-word segmentation (in this paper, we use the name BPE to refer specifically to BPE for sub-word segmentation). BPE detects repeated patterns in the text and compresses them into one sub-word. Specifically, it initializes a vocabulary of all types of characters in the training corpora, and adds frequent fragments and words into it. During decoding, a greedy algorithm recursively combines the most frequent adjacent fragment pair in the vocabulary, starting from words that are split into characters. Although not linguistically motivated, the effectiveness may come from the ability of generating shorter sequences~\cite{galle-2019-investigating}.
There are several variants of the BPE method, BPE-dropout~\cite{provilkov-etal-2020-bpe} is a stochastic or regularized version of BPE where words can be segmented in different ways causing a sentence to have multiple-segmented forms leading to a robust translation model. Subword regularization~\cite{1804.10959} is a regularized version of SentencePiece~\cite{kudo-2018-subword} based on a non-neural network unigram language model.
VOLT~\cite{xu-etal-2021-vocabulary} finds the best BPE token dictionary with a proper size. Byte-level BPE (BBPE)~\cite{1909.03341} uses bytes as the minimal unit, thus generating a compact vocabulary. WordPiece (WPM)~\cite{6289079} is similar to BPE where it chooses the adjacent fragment pair that maximizes the likelihood of the training data rather than based on word frequency. Different from BPE which treats space as a special token and thus needs a tokenizer for data in different languages, SentencePiece (SPM)~\cite{kudo-richardson-2018-sentencepiece} is a language-independent method that treats the input as a raw input stream where space is not a special token. SentencePiece regularization~\cite{kudo-2018-subword} is the stochastic version of SPM where it draws multiple segmentations from one sentence to improve the robustness of the model.

The frequency-based methods however are not linguistically motivated, for example, the word ``moments'' will be segmented as ``mom+ents'' rather than ``moment+s''. Attempts to use a morphological analyzer for sub-word segmentation cannot achieve consistently translation quality improvements \cite{giulio2018, huck-etal-2017-target}. Furthermore, this method cannot be applied to low-resource languages which lack high-quality morphological analyzers. 
A recent survey~\cite{2112.10508} also covers other non-neural methods such as language-specific methods~\cite{koehn-knight-2003-empirical}, bayesian
language models~\cite{teh-2006-hierarchical}, and marginalization over multiple possible
segmentations~\cite{1610.03035}. 


\subsection{Neural Sub-word Segmentation}

Frequency-based methods, such as BPE and SPM, are simple forms of data compression~\cite{gage1994new} to reduce entropy, which makes the corpus easy to learn and predict~\cite{1606.04289}. While we can optimize the choice of vocabulary to further reduce the entropy~\cite{xu-etal-2021-vocabulary}, it is more straightforward to find the segmentation that directly reduces the entropy of a neural model. 

Segmentations can be optimized for a neural model to learn and generate by the \textit{sequence modeling via segmentations} method~\cite{1702.07463}. In the training phase, it optimizes the sequence generation probability calculated by the sum of probabilities of all its possible segmentations. In the decoding phase, the segmentation with maximum a posteriori (MAP) is considered the optimal segmentation for each sentence. 
The sequence modeling via segmentation idea is applied to multiple NLP tasks including word segmentation~\cite{kawakami-etal-2019-learning, sun-deng-2018-unsupervised, 2104.07829},  language modeling~\cite{grave-etal-2019-training},  NMT~\cite{1810.01480}, and speech recognition~\cite{1702.07463}. During the inference of the language model, utilizing the marginal likelihood with multiple segmentations shows more robust results than one-best-segmentation~\cite{2109.02550}.
DPE~\cite{he-etal-2020-dynamic} method has applied this sequence modeling and optimization idea to the sub-word segmentation task. They proposed a mixed character-sub-word transformer and apply the dynamic programming (DP) algorithm to accelerate the calculation of sequence modeling.
However, segmentation is performed at the sentence-level and conditioned on a sentence in another language. DPE's parallel corpus requirement makes it unattractive, especially in low-resource settings, which motivated us to rely only on monolingual corpora. However, the mixed character-sub-word transformer is indispensable to our method.

\subsection{Self-supervised Machine Learning}
Self-supervised methods are becoming popular in machine learning. The advantage of this approach is that it requires only unlabeled (and often monolingual) data, which exists in large quantities.
In the NLP field, using monolingual data with denoising objectives has led to significant performance gains in multiple tasks including NMT, question answering (QA) and Multi-Genre Natural Language Inference (MultiNLI) tasks \cite{devlin-etal-2019-bert, 2019arXiv190502450S, 1910.10683, 2005.14165, 1907.11692, 10.1162/tacl_a_00343, lewis-etal-2020-bart}. However, to the best of our knowledge, this approach has not been seriously applied to the sub-word segmentation task yet.
Furthermore, the self-supervised method is prevalent in the field of computer vision. There are many works that use unlabeled images to pre-train models \cite{Hinton504, 10.1145/1390156.1390294, 1604.07379, Doersch_2015_ICCV, zhang2016colorful, 1603.08561, Wei_2018_CVPR, Vondrick_2018_ECCV}.
\section{Methods}
We first describe the sequence modeling via segmentation for the sub-word segmentation task as background in Section~\ref{word_modeling}. We then describe the proposed segmenter with several masking strategies in Section~\ref{selfseg}, word frequency normalization strategies to accelerate the training speed in Section~\ref{frequency_norm}, and a regularization mechanism to increase the variety of the generated sub-words in Section~\ref{regularization}.

\subsection{Background: Word Modeling via Sub-word Segmentations}
\label{word_modeling}

This section describes the word modeling via sub-word segmentation, which is the theoretical foundation of the proposed method. 


Let $\bm{x}_{1:T}$ denote a word that comprises $T$ characters, that is $\bm{x}_{1:T}=(x_1, ..., x_T)$. Let $\bm{a}_{1:\tau_a}$ denote one segmentation of $\bm{x}_{1:T}$ that comprises $\tau_a$ sub-words, that is $\bm{a}_{1:\tau_a}=(a_1, ..., a_{\tau_a})$. For each sub-word (or segment) $a_i$ in a segmentation $\bm{a}_{1:\tau_a}$, it is non-empty substrings of $\bm{x}_{1:T}$ and in a predefined finite size sub-word vocabulary $V$, that is $a_i \in V$. The set of all valid segmentations for a word is represented as $S_x$, where $\forall \bm{a}, \bm{a} \in S_x$. 
Because the sub-word segmentation of one word is not known in advance, the probability of generating one word $\bm{x}_{1:T}$ can be defined as the sum of the probability from all sub-word segmentations in $S_x$:

\begin{align}
\begin{split}
    p(\bm{x}_{1:T}) &= \sum_{\bm{a}_{1:\tau_a} \in S_{\bm{x}}} p(\bm{a}_{1:\tau_a}) \\
         &= \sum_{\bm{a}_{1:\tau_a} \in S_{\bm{x}}} \prod_{i=1}^{\tau_a} p(a_i|a_1, ..., a_{i-1}),
\end{split}
\label{eq:word_prob_1}
\end{align}
where $p(\bm{x}_{1:T})$ is the probability of the word, $p(\bm{a}_{1:\tau_a})$ is the probability of one segmentation and $p(a_i|a_1, ..., a_{i-1})$ is the probability of one segment in the segmentation $\bm{a}_{1:\tau_a}$, conditioned on previous segments, which is calculated using neural networks such as RNN or Transformer models.

However, for a sequence of length $T$, there are approximately $2^T$ types of segmentations. Without using approximation algorithms the time complexity of calculating \Eq{word_prob_1} will be exponential (O($2^T$)), which makes the algorithm too slow thus impractical. To address this, we adopt the mixed character-sub-word transformer model \cite{he-etal-2020-dynamic} which takes characters as input and generates sub-words as output. The model represent the history information by prefix characters $x_1, ..., x_j$ instead of sub-words $a_1, ..., a_{i-1}$, where $j=index(a_i)-1$. Therefore, we have an approximate word probability:
\begin{align}
\begin{split}
    p(\bm{x}_{1:T}) = \sum_{\bm{a}_{1:\tau_a} \in S_{\bm{x}}} \prod_{i=1}^{\tau_a} p(a_i|x_1, ..., x_j)
\end{split}
\label{eq:word_prob_2}
\end{align}

In this way, we can calculate the word probability in the time complexity $O(T^2)$, because there are only $T$ types prefixes as history states, from $x_1$, $x_1x_2$ to $x_1...x_T$, and only maximum $T$ types of possible next segments from $x_i$, $x_ix_{i+1}$ to $x_i ...x_T$, suppose the current index is $i$. This is a DP algorithm and helps speed up the segmentation process.

In the training phase, the generation probability of the model for the unsegmented sequences is optimized. 
\Fig{sub-word_segmentation_train_test} provides an example. During the training phase, we can obtain the probability of the word ``watching'' by summing the probabilities of all possible sub-word segmentations such as ``watch+ing'' and ``wat+ching,'' where the probability of each segmentation is the product of the probability of all its segments following the chain rule, calculated by a neural model. 
The training objective for this unsupervised task is to maximize the generation probability of all words: $\sum_{\bm{x}_{1:T} \in D}\log P(\bm{x}_{1:T})$ where $D$ is the training corpus consisting of the words. For one word $x$ the marginalization $P(\bm{x}_{1:T})$ is the sum of probabilities of all possible segmentations, calculated through \Eq{word_prob_2}.
The gradient is calculated automatically through PyTorch and then propagated. The detailed calculation process can be found in Section 3.1 of the sequence modeling work~\cite{1702.07463}.
In the decoding phase, we calculate the probabilities of all segmentations and then trace the one with maximum probability as the optimal segmentation.  

\subsection{SelfSeg: Self-supervised Sub-word Segmentation Method} 
\label{selfseg}

We propose a self-supervised method to train a sub-word segmenter. Given a masked version of one word, the segmenter maximizes the likelihood of all segmentations of the word during training, and selects one segmentation with the highest likelihood during decoding.

The masked version of the word is denoted by $\bm{x}_M$. And we maximize the generation probability of word $\bm{x}_{1:T}$ during training by the following objective:

\begin{align}
\begin{split}
    \log p(\bm{x}_{1:T}|\bm{x}_M) = \log  \sum_{\bm{a}_{1:\tau_a} \in S_{\bm{x}}} \prod_{i=1}^{\tau_a} p(a_i|\bm{x}_M, x_1, ..., x_j)
\end{split}
\label{eq:selfseg_word_modeling}
\end{align}

\noindent We propose the \textbf{charMASS} to generate $\bm{x}_M$:
\begin{itemize}
    \item \textbf{charMASS}: character-level MAsked Sequence-to-Sequence pre-training (charMASS), where half of the consecutive characters in one word are masked. We select the start position of the span from the indexes of the first half of the characters.
\end{itemize}
In addition, we consider three alternatives:
\begin{itemize}
    \item \textbf{subwordMASS}: sub-word level MAsked Sequence-to-Sequence pre-training (MASS), where half of the consecutive sub-word segments in one word are masked. We select the start position of the span from the indexes of the first half of the sub-words.
    \item \textbf{subwordMASK}: strategy used in the MASKed language model, where every sub-word segment is individually masked with a certain probability. We set it to $15\%$ following the BERT paper~\cite{devlin-etal-2019-bert}.
    \item \textbf{w/o masking}: where we set $\bm{x}_M$ to the original word $\bm{x}$ without any masks.
\end{itemize}

\begin{figure}[htb]
\centering  
\includegraphics[width=0.72\columnwidth]{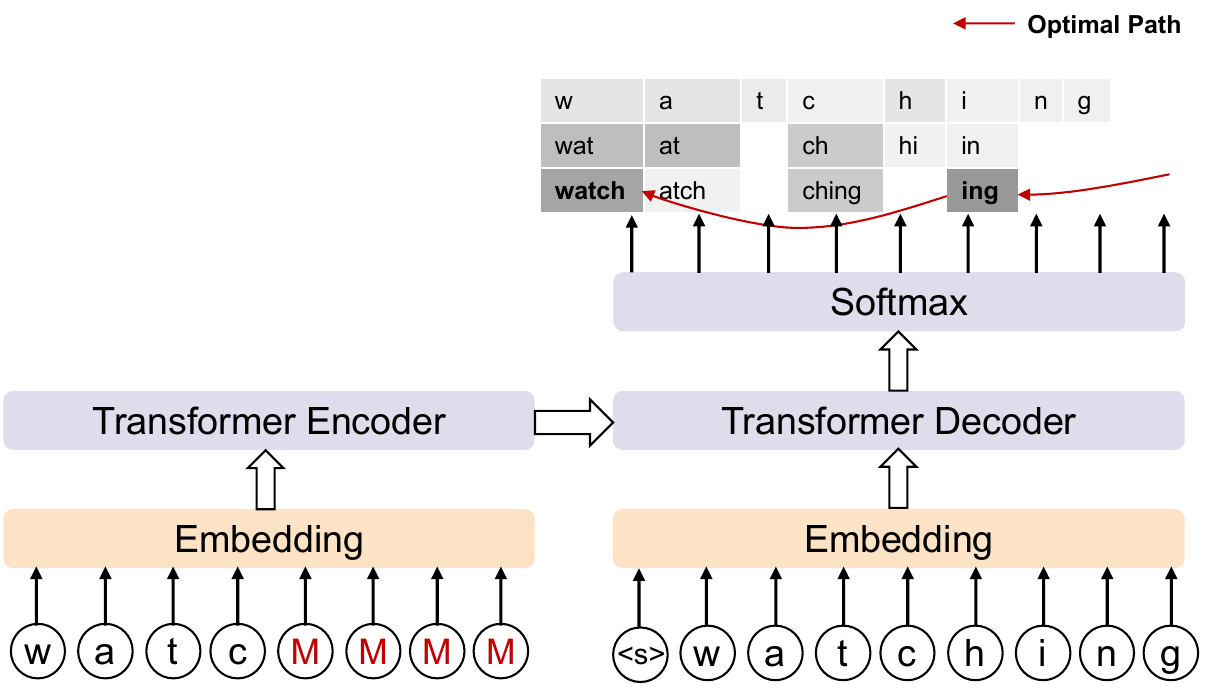}
\caption{\textbf{Mixed character-subword transformer.} The input of the encoder is one word with masks. The output of the decoder is the possibilities of the next sub-words in each position. We optimize all paths during training and retrace the optimal path during decoding.}
\label{fig:self_supervised_model}
\end{figure}

 \Fig{self_supervised_model} illustrates the charMASS method. We directly mask characters in charMASS. However, we generate an initial segmentation using existing sub-word segmentation methods such as BPE~\cite{sennrich-etal-2016-neural}, and mask part of the sub-words. We generate the next sub-word possibilities for each position. The training objective is to maximize the possibility of all paths and in the decoding phase we retrace the optimal path.
We create the word-level data by splitting sentence-level data into one word per line format. During decoding, we decode each type of word once which accelerates the decoding phase.

\subsection{Word Frequency Normalization} 
\label{frequency_norm}
We propose frequency normalization methods to speed up the training phase.
The motivation is the observation that high-frequency words make up a large part of the training set, such as the words ``the'' and ``is''. However, they can not provide sufficient training signals because most of them are short and non-compound words and tend to stay unsegmented.

Suppose word $w_i$ occures $q_i$ times in the corpus. And $freq$ is a function that maps $w_i$ into $q_i$. We propose normalizing function $norm$ acting on the function $freq$ and generate normalized frequency $nq_i$ for each word, that is $norm\circ freq(w_i)=nq_i$.

\noindent We propose the \textbf{Threshold} as $norm$
\begin{itemize}
    \item \textbf{Threshold}: $threshold(x)=\lfloor x/d \rfloor$, where we remove words with frequency lower than a threshold $d$ and reduce the frequency for other words. We set $d$ to 10.
\end{itemize}
In addition, we consider three alternatives:
\begin{itemize}
    \item \textbf{Sqrt}: $sqrt(x)=\lfloor \sqrt{x} \rfloor$, in this way we reserve all types of words while especially reduce the frequency of high-frequency words. 
    \item \textbf{Log}:   $log(x)=\lfloor \log_2 x \rfloor$, where we also reserve all types of words and cuts the frequency of high-frequency words more strongly. 
    \item \textbf{One}:  $One(x)=1$, where we retain only the type information and removes the frequency information.
\end{itemize}

We create the training data by 1) obtaining a word-frequency table from the corpus, 2) applying the normalizing function and obtaining the normalized word-frequency table, and 3) copying each word $w_i$ by $nq_i$ times and then shuffle the dataset.

\subsection{SelfSeg Regularization}
\label{regularization}
\Algo{selfseg_regularization} shows the proposed SelfSeg-Regularization algorithm that is to increase the variety of the generated sub-words.
At each position $i$ of word $\bm{x}$ during decoding, we calculate the scores $\beta_{j}$ of choosing the sub-word $\bm{x}_{j:i}$.
Instead of selecting the index $j$ with the highest score, we perform weighted random sampling to draw the next sub-word.
As shown in Line $5$, the weights are calculated by feeding the probability of each index $j$ to a softmax function with temperature $t$ to control the diversity. We save the $idx_i$ and retrace the segmentation $\bm{z}$ for each run. During decoding, for each type of word, we run the algorithm $N$ times to generate a list of $N$ segmentations.

\begin{algorithm}
    \caption{SelfSeg-Regularization}
    \label{algo:selfseg_regularization}
    \KwIn{$\bm{x}$ is a word containing $T$ characters, $V$ is a sub-word vocabulary, $t$ is the temperature hyperparameter.}
    \KwOut{Segmentation $\bm{z}$}
    \For{$i \gets 1$ \textbf{to} $T$}
    {
        \For{$j \gets 1$ \textbf{to} $i$}
        {
            \uIf{$\bm{x}_{j:i} \in V$}
            {
                $\beta_{j} \gets \alpha_{j-1}+\log P(\bm{x}_{j:i}|x_1, .., x_{j-1})$\;
            }
        }
        $idx_{i} \gets randomChoice([1, ..., i], weights=softmax(\beta/t))$\;
        $\alpha_{i} \gets \beta_{idx_{i}}$\;
    }
    $\bm{z} \gets retrace(\bm{idx})$\;
    \Return{$\bm{z}$}\;
\end{algorithm}

\section{Experimental Settings}
\label{exp_settings}
\subsection{Datasets}

We experimented with low-resource, middle-resource, and high-resource MT settings. The datasets are listed in~\Tab{datasets}, where the size of the vocabulary is set for both the segmenters\fn{We keep in line with SPM's definition of vocabulary size.} and NMT models for all methods, if not otherwise specified. We applied Juman++ \citep{tolmachev-etal-2018-juman} for Japanese, Stanford-tokenizer \citep{manning-etal-2014-stanford} for Chinese and Moses tokenizer \citep{koehn-etal-2007-moses} to data of all the other languages.
 We normalized Romanian data and removed diacritics following previous work~\cite{sennrich-etal-2016-edinburgh}. 

\begin{table}[htb]
    \caption{Statistics of the corpora used in the NMT experiments.}
    \centering
    \begin{tabu}{c|r|r|r|r}
    \toprule
   \textbf{Dataset} & \textbf{Train} & \textbf{Valid} & \textbf{Test} & \textbf{Vocab} \\
    \midrule
    ALT Asian Langs-En & $18k$ & $1,000$ & $1,018$ & $8k$ \\
    IWSLT15 Vi-En & $133k$ & $1,553$ & $1,268$ & $8k$ \\
    WMT16 Ro-En & $612k$ & $1,999$ & $1,999$ & $8k$ \\
    WMT15 Fi-En & $1.8M$ & $1,500$ & $1,370$ & $8k$ \\
    WMT14 De-En & $4.5M$ & $45,781$ & $3,003$ & $8k$ \\
    \rowfont{\color{blue}}
    WMT14 Fr-En & $10.0M$ & $26,875$ & $3,003$ & $8k$ \\
    \bottomrule
    \end{tabu}
    \label{tab:datasets}
\end{table}

\noindent

\textbf{Low-resource Setting} We used the ALT multi-way parallel dataset \citep{thu-etal-2016-introducing}. We used English and $6$ Asian languages: Filipino (Fil), Indonesian (Id), Japanese (Ja), Malay (Ms), Vietnamese (Vi), and simplified Chinese (Zh).
The SelfSeg segmenter is applied to only the target language side. Therefore, we train one SelfSeg segmenter using $50,000$ randomly selected English sentences from news commentary corpus\fn{\url{http://data.statmt.org/news-commentary/v14/}} for all the Asian language to English directions. We trained a Japanese SelfSeg segmenter using $98k$ Japanese sentences from KFTT dataset~\cite{neubig11kftt} for English to Japanese direction and an Indonesian SelfSeg segmenter using $62k$ Indonesian sentences from the Indonesian news commentary corpus for English to Indonesian direction. We trained one DPE~\citep{he-etal-2020-dynamic} segmenter for each language pair in ALT using the corresponding $18,088$ parallel sentences. We trained BPE~\cite{sennrich-etal-2016-neural}, BPE-dropout~\cite{provilkov-etal-2020-bpe} and VOLT~\cite{xu-etal-2021-vocabulary} segmenters using the $18,088$ monolingual sentences in ALT for the corresponding languages. 

\noindent

\textbf{Middle- and High- Resource Setting} We used the IWSLT'15 Vietnamese-English, WMT'16 Romanian-English, WMT'15 Finnish-English, WMT'14 German-English,\fn{\url{https://github.com/facebookresearch/fairseq/blob/main/examples/translation/prepare-wmt14en2de.sh}} and \hl{WMT'14 French-English\fn{\url{https://github.com/facebookresearch/fairseq/blob/main/examples/translation/prepare-wmt14en2fr.sh}} corpora. We use the first $10.0$ million parallel sentence pairs in the WMT'14 French-English train set in our experiments.} We used English monolingual sentences from the training set of each corpus as the training data for all methods except DPE. For the DPE method, we used the parallel sentences from the train sets \hl{following the official implementation,\fn{\url{https://github.com/xlhex/dpe}} where the input of the encoder is the sentence in the source language, and the predicted output is the sentence in the target language.}

\subsection{Segmenter Model Settings}
\label{segmenter_settings}

\noindent

\textbf{BPE, SentencePiece, VOLT, and BPE-dropout}
For the BPE~\cite{sennrich-etal-2016-neural} method, we used a widely adopted toolkit\fn{\url{https://github.com/google/sentencepiece}} with model type as BPE. For SentencePiece, we use unigram language model implemented in the toolkit. For VOLT~\cite{xu-etal-2021-vocabulary}, we used the default setting in the official implementation.\fn{\url{https://github.com/Jingjing-NLP/VOLT}} For BPE-dropout~\cite{provilkov-etal-2020-bpe}, we apply dynamic dropout for each epoch and with a drop rate of $0.1$ ($0.05$ for English\ra Japanese) selected by hyperparameter tunning.

\noindent

\textbf{SelfSeg and DPE} For SelfSeg, we used charMASS as the masking strategy and Threshold as the word frequency normalization strategy in Section~\ref{section_result}. Detailed analysis of the masking strategies and frequency normalization strategies are shown in Section~\ref{section_analysis}. For the SelfSeg and DPE, we used the mixed character-sub-word transformer model with DP algorithm, where the transformer architecture is of $4$ encoder layers and $4$ decoder layers, dropout of $0.3$, inverse sqrt learning rate scheduler with $4,000$ warmup steps, and the dynamic programming cross-entropy criterion as described in the DPE method. We set the number of training epochs to $50$, which is large enough for convergence.
Additionally, for the mixed character-sub-word transformer model, the vocabulary $V$ should contain all characters to prevent OOV problems and commonly used sub-words. Here we used a sub-word vocabulary generated by the BPE algorithm~\cite{sennrich-etal-2016-neural}, which satisfies the two conditions, following previous work~\cite{he-etal-2020-dynamic}.

\noindent

\textbf{SelfSeg-Regularization} We set $N$ to 10 and $t$ to $10$ ($t$ to $3$ for English\ra Japanese) in \Algo{selfseg_regularization}. In the MT experiments, at each epoch, we dynamically generate a segmentation for each sentence in the dataset. For each word in the sentence, we randomly select one of the $N$ segmentations.

Note that DPE, SelfSeg, and SelfSeg-Regularization are used to segment only the target side in the MT experiments. The source-side simply uses BPE data for SelfSeg and BPE-dropout data for the SelfSeg-Regularization. This is because the loss function of the segmenter is to maximize the generation probability. Therefore, these segmentations are effective for the target sentence. This is also studied in the DPE work~\cite{he-etal-2020-dynamic}.

\subsection{NMT Settings}
We used the fairseq framework \citep{ott-etal-2019-fairseq} with the Transformer \citep{NIPS2017_3f5ee243} architecture with $6$ layer encoder (except for Filipino where $4$ encoder layers were sufficient), $6$ layer decoder and $1$ attention head, decided through hyperparameter tuning as suggested by \citet{DBLP:journals/mt/RubinoMDFUS20}. Dropout of $0.1$ and label smoothing of $0.1$ is used. We used layer normalization \citep{2016arXiv160706450L} for both the encoder and decoder.
We used a vocabulary size of $8,000$ for the NMT models.
Batch-size is set to $1,024$ tokens. We used the
ADAM optimizer \citep{2014arXiv1412.6980K} with betas ($0.9$, $0.98$), warm-up of $4,000$ steps followed by decay, and performed early stopping based on the validation set BLEU.   
We used a beam size of $12$ and a length penalty of $1.4$ for decoding. We reported sacreBLEU~\citep{post-2018-call}, METEOR~\cite{banerjee-lavie-2005-meteor}, and BLEURT~\cite{sellam-etal-2020-bleurt} on detokenized outputs.

\section{Results}
\label{section_result}
We report the performance of NMT as well as the training/decoding speed of our methods compared with existing works in this section.

\subsection{MT Results}
\noindent{\textbf{Low-Resource Scenario}} 
Tables~\ref{tab:low-resource-res} and~\ref{tab:low-resource-res-other-metrics} show low-resource Asian language to English NMT results. 
SelfSeg-Regularization achieves the highest BLEU scores among all methods in almost all directions, outperforming the BPE method by $4.31$ BLEU scores on average.
Among methods without regularization, proposed SelfSeg outperforms not only frequency-based methods but also neural method DPE. However, we observed that for the Ms\ra En and Zh\ra En directions, the proposed SelfSeg method is slightly worse (which is not significant) than the BPE method. In particular, we find that both neural methods (DPE and SelfSeg) perform relatively poorly in the Zh\ra En direction. Actually, for all directions SelfSeg are better (or worse) than BPE, DPE is also better (or worse) than BPE. Therefore, we assume that for segmentations generated by neural segmenters, the performance do have a correlation with the source language. We will leave the in-depth exploration of this question as future work.
We found that adding regularization yields significant BLEU score improvement in the low-resource situation. The SelfSeg-Regularization method substantially improves over BPE. Results of the METEOR and BLEURT evaluation metrics also show similar trends.

Tables~\ref{tab:low-resource-res2} and~\ref{tab:low-resource-res2-other-metrics} show English to Japanese and Indonesian NMT results \hl{of the ALT dataset, English to Romanian results of the WMT16 Ro-En dataset, and the English to Finnish results of the WMT15 Fi-En dataset}. In English to Indonesian direction, the SelfSeg-Regularization outperforms all baseline methods substantially.
For the English to Japanese direction, the improvement is limited because the average length of the Japanese words in the ALT dataset is short, only $1.87$, resulting in less variety in word segmentation. As a comparison, the average length of English words is $4.54$ and the average length of Indonesian words is $5.50$. This may explain why regularization brings more improvement for English\ra Indonesian than English\ra Japanese. \hl{For the En\ra Ro and En\ra Fi translation directions, we observed that the SelfSeg performs best among the w/o regularization methods whereas the results of BPE-dropout and SelfSeg-regularization are comparable in terms of the BLEU, METEOR and BLEURT metrics.}

\renewcommand{\arraystretch}{1.3}
\begin{table}[thb]
    \caption{\textbf{Low-resource Asian languages to English MT results.} The numbers in the table indicate the sacreBLEU scores. We show the average BLEU scores (Avg) and the improvements ($\Delta$) over the BPE method. Methods are separated into without regularization and with regularization.  \hl{Statistical significance~\cite{koehn-2004-statistical} is indicated by $\superscript{*}$ $(p < 0.001)$ between the BPE baseline and the proposed methods in each direction.}}
    \resizebox{0.85\textwidth}{!}{
    \begin{tabu}{cSSSSSS|SS}
        \toprule
         & \textbf{Fil}\ra \textbf{En} & \textbf{Id}\ra \textbf{En} & \textbf{Ja}\ra \textbf{En} & \textbf{Ms}\ra \textbf{En} & \textbf{Vi}\ra \textbf{En} & \textbf{Zh}\ra \textbf{En} & \textbf{Avg} & \textbf{$\Delta$} \\ 
        \toprule
        \textit{w/o Regularization} &&&&&&&& \\
       
        BPE~\cite{sennrich-etal-2016-neural} & 23.09&25.70&9.42& 28.19&19.94& 12.21&19.76&0.00 \\
       
        SentencePiece~\cite{kudo-2018-subword}&23.71&25.49&9.94&27.72&18.58&11.74&19.53& -0.23\\
       
        VOLT~\cite{xu-etal-2021-vocabulary}&22.99&25.05&10.56&27.91&21.64&11.31&19.91&0.15 \\
       
        DPE~\citep{he-etal-2020-dynamic}&24.04&26.66&9.93&27.89&20.06&10.72&19.88&0.13\\
       
        \bfseries{SelfSeg}&  25.20 \superscript{*}& 27.10\superscript{*}& 11.39\superscript{*}& 28.15& 22.44\superscript{*}&12.03& 21.05& 1.29\\
        \toprule
       
        \textit{With Regularization} &&&&&&&& \\
       
        BPE-dropout~\cite{provilkov-etal-2020-bpe}&28.18&28.02&12.84&31.59&23.67&\textbf{13.91}&23.04&3.13\\
       
        \bfseries SelfSeg-Regularization&\bfseries 29.94\superscript{*}&\bfseries 29.34\superscript{*}&\bfseries 15.23\superscript{*}&\bfseries 32.31\superscript{*}&\bfseries 23.93\superscript{*}&13.64\superscript{*}&\bfseries 24.07&\bfseries 4.31\\
        \bottomrule
    \end{tabu}
    }
    \label{tab:low-resource-res}
\end{table}

\begin{table}[thb]
    \caption{\textbf{Low-resource Asian languages to English MT results.} The numbers in the table indicate the METEOR~\cite{banerjee-lavie-2005-meteor}/BLEURT~\cite{sellam-etal-2020-bleurt} scores. We show the average scores and the improvements ($\Delta$) over the BPE method.}
    \resizebox{\textwidth}{!}{
    \begin{tabu}{cSSSSSS|SS}
        \toprule
         & \textbf{Fil}\ra \textbf{En} & \textbf{Id}\ra \textbf{En} & \textbf{Ja}\ra \textbf{En} & \textbf{Ms}\ra \textbf{En} & \textbf{Vi}\ra \textbf{En} & \textbf{Zh}\ra \textbf{En} & \textbf{Avg} & \textbf{$\Delta$} \\ 
        \toprule
        \textit{w/o Regularization} &&&&&&&& \\
        BPE~\cite{sennrich-etal-2016-neural} & \footnotesize{29.1/45.0}&\footnotesize{31.1/49.2}&\footnotesize{20.1/32.4}&\footnotesize{32.7/52.0}&\footnotesize{27.6/44.6}&\footnotesize{22.9/36.9}&\footnotesize{27.2/43.3}&\footnotesize{0.0/0.0}\\
        SentencePiece~\cite{kudo-2018-subword}&\footnotesize{29.7/46.1}&\footnotesize{31.2/48.9}&\footnotesize{21.0/33.8}&\footnotesize{32.2/51.0}&\footnotesize{26.6/42.4}&\footnotesize{21.6/34.2}&\footnotesize{27.0/42.7}&\footnotesize{-0.2/-0.6}\\
        VOLT~\cite{xu-etal-2021-vocabulary}&\footnotesize{29.2/45.2}&\footnotesize{31.0/48.8}&\footnotesize{21.2/34.2}&\footnotesize{32.5/51.1}&\footnotesize{28.4/46.6}&\footnotesize{22.2/35.5}&\footnotesize{27.4/43.6}&\footnotesize{0.2/0.2}\\
        DPE~\citep{he-etal-2020-dynamic}&\footnotesize{29.7/46.5}&\footnotesize{31.8/50.5}&\footnotesize{21.1/34.4}&\footnotesize{32.5/51.6}&\footnotesize{26.9/43.9}&\footnotesize{21.5/35.3}&\footnotesize{27.3/43.7}&\footnotesize{0.0/0.3}\\
        \bfseries{SelfSeg}&\footnotesize{30.2/47.3}&\footnotesize{32.0/51.3}&\footnotesize{21.5/35.3}&\footnotesize{32.6/52.3}&\footnotesize{28.4/46.3}&\footnotesize{22.4/36.5}&\footnotesize{27.9/44.8}&\footnotesize{0.6/1.5}\\
        \toprule
        \textit{With Regularization} &&&&&&&& \\
        BPE-dropout~\cite{provilkov-etal-2020-bpe}&\footnotesize{32.0/51.1}&\footnotesize{33.0/52.2}&\footnotesize{22.8/36.9}&\footnotesize{34.8/55.8}&\footnotesize{29.1/48.3}&\footnotesize{\textbf{23.6}/38.8}&\footnotesize{29.2/47.2}&\footnotesize{2.0/3.8}\\
        \bfseries SelfSeg-Regularization&\footnotesize{\textbf{33.2/52.6}}&\footnotesize{\textbf{33.5/53.9}}&\footnotesize{\textbf{24.4/40.1}}&\footnotesize{\textbf{35.0/56.5}}&\footnotesize{\textbf{29.7/48.7}}&\footnotesize{23.0/\textbf{38.8}}&\footnotesize{\textbf{29.8/48.4}}&\footnotesize{\textbf{2.6/5.1}}\\
        \bottomrule
    \end{tabu}
    }
    \label{tab:low-resource-res-other-metrics}
\end{table}

\renewcommand{\arraystretch}{1.3}
\begin{table}[thb]
    \small
    \caption{\hl{En\ra Other language results. En\ra Ja and En\ra Id directions are from the ALT dataset, En\ra Ro direction is from the WMT16 Ro-En dataset and En\ra Fi direction is from the WMT15 Fi-En dataset.}  \hl{Statistical significance~\cite{koehn-2004-statistical} is indicated by $\superscript{*}$ $(p < 0.001)$ between the BPE baseline and the proposed methods in each direction.}}
    \begin{tabu}{ccccc}
        \toprule
        & \textbf{En}\ra \textbf{Ja} & \textbf{En}\ra \textbf{Id} & \textbf{\hl{En}}\ra \textbf{\hl{Ro}} & \textbf{\hl{En}}\ra \textbf{\hl{Fi}}\\
        \midrule
        \textit{w/o Regularization} &&\\
        BPE~\cite{sennrich-etal-2016-neural} & 12.69 & 28.08 & \hl{33.62} & \hl{15.54} \\
        SentencePiece~\cite{kudo-2018-subword} & 12.58 & 26.01 & \hl{33.17} & \hl{15.75} \\
        VOLT~\cite{xu-etal-2021-vocabulary} & 13.11 & 28.46 & \hl{33.13} & \hl{15.24} \\
        DPE~\citep{he-etal-2020-dynamic}& 13.46 & 29.29 & \hl{33.71} & \hl{15.27} \\
        \textbf{SelfSeg} & 13.26 & 29.00 &\hl{33.72} & \hl{15.85} \\
        \toprule
        \textit{With Regularization} &  & \\
        BPE-dropout~\cite{provilkov-etal-2020-bpe} & \bfseries 14.97  & 30.74 & \hl{\textbf{35.48}} & \hl{\textbf{17.04}} \\
        \textbf{SelfSeg-Regularization} & 14.31\superscript{*} & \bfseries 33.77\superscript{*} & \hl{35.47\superscript{*}} & \hl{16.93\superscript{*}} \\
        \bottomrule
    \end{tabu}
    \label{tab:low-resource-res2}
\end{table}

\renewcommand{\arraystretch}{1.3}
\begin{table}[thb]
    \small
    \caption{\hl{En\ra Other language MT results. En\ra Ja and En\ra Id directions are from the ALT dataset, En\ra Ro direction is from the WMT16 Ro-En dataset and En\ra Fi direction is from the WMT15 Fi-En dataset.} The numbers in the table indicate the METEOR~\cite{banerjee-lavie-2005-meteor}/BLEURT~\cite{sellam-etal-2020-bleurt} scores.}
    \begin{tabu}{ccccc}
        \toprule
        & \textbf{En}\ra \textbf{Ja} & \textbf{En}\ra \textbf{Id}& \textbf{\hl{En}}\ra \textbf{\hl{Ro}} & \textbf{\hl{En}}\ra \textbf{\hl{Fi}} \\
        \midrule
        \textit{w/o Regularization} &&&&\\
        BPE~\cite{sennrich-etal-2016-neural} &24.87/17.94 & 30.13/46.66 &  \hl{31.19/66.92}&\hl{19.84/62.02}\\
        SentencePiece~\cite{kudo-2018-subword} &23.91/17.50 & 29.29/45.99 &\hl{31.20/66.38}&\hl{20.31/63.16}\\
        VOLT~\cite{xu-etal-2021-vocabulary} &25.10/18.45 & 30.31/46.70&    \hl{31.12/66.02}&\hl{19.79/61.86} \\
        DPE~\citep{he-etal-2020-dynamic}&25.24/18.39 & 30.76/48.00 &       \hl{31.42/66.69}&\hl{19.94/62.86}\\
        \textbf{SelfSeg} &25.24/18.45 & 30.59/47.97 &                      \hl{31.08/67.30}&\hl{19.98/62.71}\\
        \toprule
        \textit{With Regularization} &  & \\
        BPE-dropout~\cite{provilkov-etal-2020-bpe} & \textbf{26.00}/\textbf{20.84} &   31.56/48.66&\hl{\textbf{32.10/69.59}}&\hl{20.90/65.20}\\
        \textbf{SelfSeg-Regularization} & 25.47/20.65 & \textbf{33.07}/\textbf{50.72} &\hl{32.04/69.44}&\hl{\textbf{21.05/65.61}}\\
        \bottomrule
    \end{tabu}
    \label{tab:low-resource-res2-other-metrics}
\end{table}

\noindent

\textbf{Middle- and High-Resource Scenario}
The results for the middle- and high-resource scenarios are presented in Tables ~\ref{tab:mid-resource-res} and~\ref{tab:mid-resource-res-other-metrics}. The proposed methods show up to $1.9$ BLEU score improvement, $1.0$ METEOR score improvement and $1.4$ BLEURT score improvement compared with BPE and outperform other baseline methods for all datasets \hl{except the high-resource settings WMT14 De\ra En and WMT14 Fr\ra En}. \hl{Additionally, the neural methods (DPE and SelfSeg) outperform non-neural methods (BPE and SentencePiece) in most settings.} 

We find that the effect of subword segmentation on performance becomes marginal as the training data becomes larger. For the WMT14 De\ra En \hl{and WMT14 Fr\ra En} directions, we found no improvement over BPE. Additionally, two methods with regularization didn't show better results than methods without regularization. 
This is also shown in the DPE work~\cite{he-etal-2020-dynamic} where the improvement is marginal, and the BPE-dropout work~\cite{provilkov-etal-2020-bpe} where the dropout hurts the performance for larger datasets. Therefore, one of the limitations of our approach is the small to medium sized MT dataset. Note that we didn't conduct DPE experiments on the WMT14 De\ra En \hl{and WMT14 Fr\ra En} datasets because of excessive computational resource consumption \hl{as shown in \Sec{speed}}.

\renewcommand{\arraystretch}{1.3}
\begin{table}[thb]
    \small
    \caption{Middle- and high-resource MT results with BLEU scores. \hl{Statistical significance~\cite{koehn-2004-statistical} is indicated by $\superscript{*}$ $(p < 0.001)$ between the BPE baseline and the proposed methods in each direction.}}
    \resizebox{\textwidth}{!}{
    \begin{tabu}{cccccc}
        \toprule
        & \textbf{IWSLT15 Vi\ra En} & \textbf{WMT16 Ro\ra En} & \textbf{WMT15 Fi\ra En} & \textbf{WMT14 De\ra En} & \hl{\textbf{WMT14 Fr\ra En}}\\
        \midrule
        \textit{w/o Regularization} &&&&&\\
        BPE~\cite{sennrich-etal-2016-neural} & 27.09 & 32.54 & 17.45 & 31.00 & \hl{34.97} \\
        SentencePiece~\cite{kudo-2018-subword} & 26.58 & 31.48 & 17.74 & 30.62 & \hl{34.92} \\
        VOLT~\cite{xu-etal-2021-vocabulary} & 27.16 & 31.89 & 17.25 & \bfseries 31.24 & \hl{\textbf{35.60}}\\
        DPE~\citep{he-etal-2020-dynamic}& 27.40 & \hl{33.05} & \hl{17.51} & - & \hl{-}\\
        \textbf{SelfSeg} & 28.19 & 32.59 &  18.00 & 30.82 & \hl{34.91}\\
        \toprule
        \textit{With Regularization} &  & &&&\\
        BPE-dropout~\cite{provilkov-etal-2020-bpe} & 28.76  & 33.59 & 18.89 & 30.56&\hl{34.38}\\
        \textbf{SelfSeg-Regularization} & \: \bfseries 29.01\superscript{*}  & \:\;\bfseries 34.01\superscript{*} & \:\;\bfseries 19.01\superscript{*} & 30.59&\hl{34.39}\\
        \bottomrule
    \end{tabu}
    }
    \label{tab:mid-resource-res}
\end{table}

\renewcommand{\arraystretch}{1.3}
\begin{table}[thb]
    \small
    \caption{Middle- and high-resource MT results. The numbers in the table indicate the METEOR~\cite{banerjee-lavie-2005-meteor}/BLEURT~\cite{sellam-etal-2020-bleurt} scores.}
    \resizebox{\textwidth}{!}{
    \begin{tabu}{cccccc}
        \toprule
        & \textbf{IWSLT15 Vi\ra En} & \textbf{WMT16 Ro\ra En} & \textbf{WMT15 Fi\ra En} & \textbf{WMT14 De\ra En} & \hl{\textbf{WMT14 Fr\ra En}}\\
        \midrule
        \textit{w/o Regularization} &&&&&\\
        BPE~\cite{sennrich-etal-2016-neural} & 31.16/57.75 & 35.18/61.99 & 27.06/55.83& 34.09/64.66              & \hl{36.24/67.04}\\
        SentencePiece~\cite{kudo-2018-subword} & 30.63/56.42 & 34.43/60.64 & 27.32/56.45 & 33.49/63.68           & \hl{36.74/67.46}\\
        VOLT~\cite{xu-etal-2021-vocabulary} & 30.90/57.13 & 34.90/61.28 & 26.73/55.44 & \textbf{34.04/64.60}     & \hl{\textbf{37.00/67.80}}\\
        DPE~\citep{he-etal-2020-dynamic}& 31.07/57.61 & \hl{35.47/62.28} & \hl{27.38/55.96}& - & \hl{-}\\
        \textbf{SelfSeg} & 31.46/58.50 & 35.26/62.44 & 27.45/56.67 & 33.54/64.42 &\hl{36.17/67.31}\\
        \toprule
        \textit{With Regularization} &  & &&\\
        BPE-dropout~\cite{provilkov-etal-2020-bpe} &32.09/59.07 & 35.73/\textbf{63.38} & \textbf{28.39/58.43} & 33.59/64.18 & \hl{35.95/66.88}\\
        \textbf{SelfSeg-Regularization} &\textbf{32.15/59.17} & \textbf{35.84}/63.35& 28.11/57.87 & 33.55/63.72 & \hl{36.41/66.77}\\
        \bottomrule
    \end{tabu}
    }
    \label{tab:mid-resource-res-other-metrics}
\end{table}

\subsection{Training and Decoding Speeds} \label{sec:speed}
\begin{figure}[htb]
    \centering  
    \includegraphics[width=0.6\columnwidth]{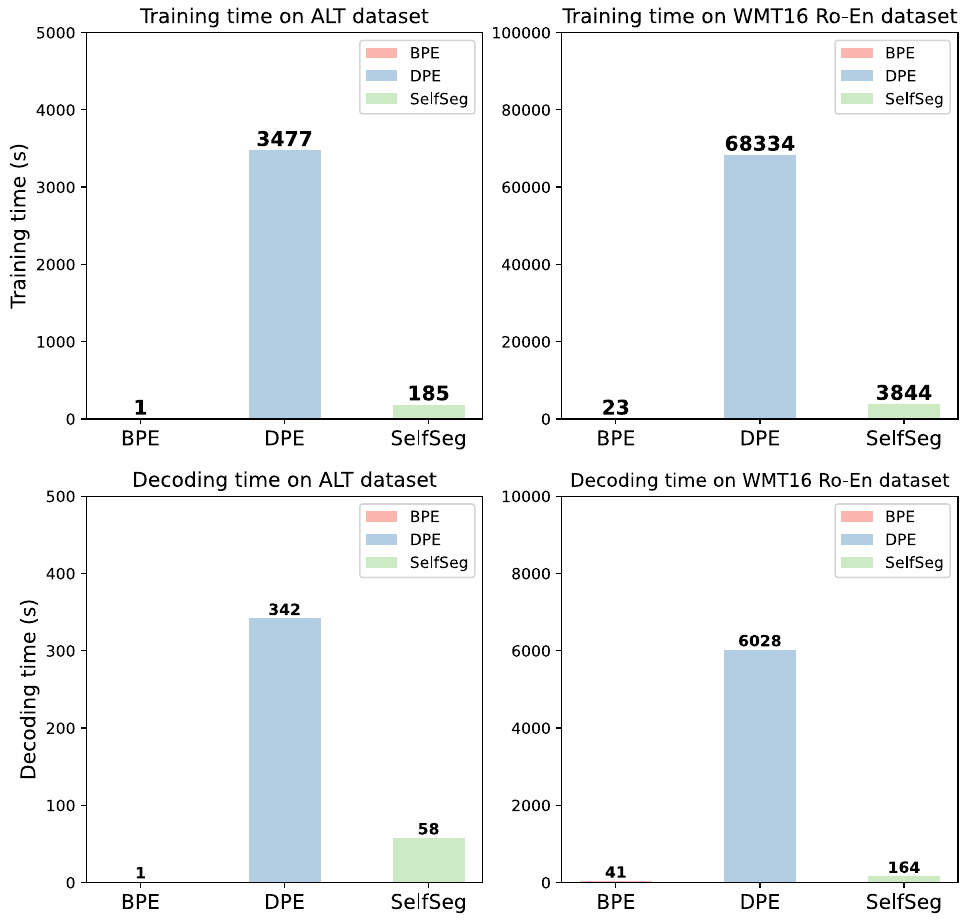}
    \caption{The training and decoding speeds of BPE, DPE, and SelfSeg methods on two datasets.}
    \label{fig:training_decoding_speed}
\end{figure}

\Fig{training_decoding_speed} provides the training speeds and decoding speeds of SelfSeg, BPE and DPE. 
The training speed of SelfSeg is $17.8$x faster than the DPE method on the WMT'16 Ro-En dataset and $18.7$x faster on the ALT dataset. Although the speed is not as fast as the BPE method, the training process can finish in approximately one hour for a $612k$ size dataset, which is much more acceptable than the DPE method which requires more than one day.

The decoding speed of SelfSeg is $5.9$x on a smaller ALT dataset and $36.8$x on a larger WMT16 Ro-En dataset compared with the DPE method. This is because, according to Zipf's law, the number of distinct words in a document increases much slower compared with the increment of the total number of words in the document, i.e $\Delta O(\#distinct\ words)\ll \Delta  O(\#total\ number\ of\ words)$. As shown in \Tab{tok_num}, for the smaller ALT dataset, DPE needs to decode $14.3$x more tokens than SelfSeg, however, for the larger WMT'16 Ro-En dataset, DPE needs to decode $186.5$x more tokens than SelfSeg. Therefore, the advantage of SelfSeg becomes greater when the corpus becomes bigger because it only needs to decode each distinct word once in the corpus.

The SelfSeg-Regularization method is only applied in the decoding phase, therefore the training time is the same as SelfSeg. During decoding, it generates $N$ segmentations for one word, therefore, the time consumption is $N$ times compared with SelfSeg. When we set $N$ to $10$, the decoding time will still be less than that of DPE.

The speed improvement is important because, in a latency-sensitive scenario, it is important to minimize as many computations as possible. Given that SelfSeg can lead to more intuitive segmentations (as seen in Section~\ref{segmentation_case_study}) and better translation than BPE while being significantly faster than DPE, which indicates that the proposed method can be very reliable in a low-latency scenario.

As a supplement, we provide statistics on how many sub-words each sentence contains. As shown in \Tab{avglength}, there is no significant difference in the number of sub-words in the sentence using different segmentation methods. For the without regularization group, the order is SentencePiece>SelfSeg>BPE>VOLT>DPE. For the with regularization group, BPE-dropout>Selfseg-regularization. This shows that the number of sub-words is not a key reason for the speed difference.

\begin{table}[htb]
\caption{The average number of sub-words in each English sentence by different segmenters.}
\resizebox{0.78\textwidth}{!}{
    \begin{tabu}{c|cccccc}
    \toprule
     & \multirow{2}{*}{\shortstack{\textbf{ALT}\\\textbf{Asian Langs\ra En}}} & \multirow{2}{*}{\shortstack{\textbf{IWSLT15}\\\textbf{Vi\ra En}}} & \multirow{2}{*}{\shortstack{\textbf{WMT16}\\\textbf{Ro\ra En}}} & \multirow{2}{*}{\shortstack{\textbf{WMT15}\\\textbf{Fi\ra En}}} & \multirow{2}{*}{\shortstack{\textbf{WMT14}\\\textbf{De\ra En}}}& \multirow{2}{*}{\shortstack{\textbf{\hl{WMT14}}\\\textbf{\hl{Fr\ra En}}}} \\
     & & & & & &\\
     \midrule
     \textit{w/o Regularization} &&&&&&\\
    BPE~\cite{sennrich-etal-2016-neural}                    & 34.04 & 24.80 & 30.40 & 26.63 & 35.41  &\hl{35.33}           \\
    SentencePiece~\cite{kudo-2018-subword}          & 41.00 & 28.15 & 35.30 & 29.39 & 35.79  &\hl{35.15}               \\
    VOLT~\cite{xu-etal-2021-vocabulary}                   & 34.04 & 25.14 & 29.60 & 26.03 & 32.88 &\hl{32.74}                \\
    DPE~\citep{he-etal-2020-dynamic}                & 34.17 & 24.62 & 27.44 & 25.67 & \multicolumn{1}{c}{-} &\hl{-} \\
   \textbf{Selfseg}                & 40.31 & 25.92 & 34.23 & 29.19 & 36.18  &\hl{36.74}               \\
    \toprule
     \textit{With Regularization} &&&&&&\\
    BPE-dropout~\cite{provilkov-etal-2020-bpe}            & 47.20 & 32.69 & 44.36 & 38.82 & 47.51  &\hl{49.29}               \\
   \textbf{Selfseg-regularization} & 44.51 & 32.18 & 43.50 & 38.00 & 46.67 &\hl{49.25}\\
    \bottomrule
    \end{tabu}
    }
    \label{tab:avglength}
\end{table}

\begin{table}[htb]
\small
    \caption{Number of tokens DPE and SelfSeg methods require to decode for each dataset.}
    \begin{center}
    \begin{tabular}{c|rr}
        \toprule
        & \textbf{ALT} & \textbf{WMT16 Ro-En} \\
        \midrule
        DPE & 478k & 16M \\
        SelfSeg & 33k & 70k \\
        \bottomrule
    \end{tabular}
    \label{tab:tok_num}
    \end{center}
\end{table}

%

\section{Analysis}
\label{section_analysis}
\subsection{Masking Strategies}
\Tab{masking-strategies} shows the performance of using different masking strategies. 
The charMASS method shows the highest performance, while the performance of subwordMASS is also higher than w/o masking, whereas subwordMASK is slightly worse than w/o masking. This is because the subwordMASK objective is not very suitable for the generation task.
Second, charMASS shows higher BLEU scores than subwordMASS. This is because the number of characters in the word is more than the number of sub-words. During training, charMASS can generate more variants of the masked source inputs, which provides more training signals. 

\begin{figure}[htb] 
    \centering  
    \includegraphics[width=0.48\columnwidth]{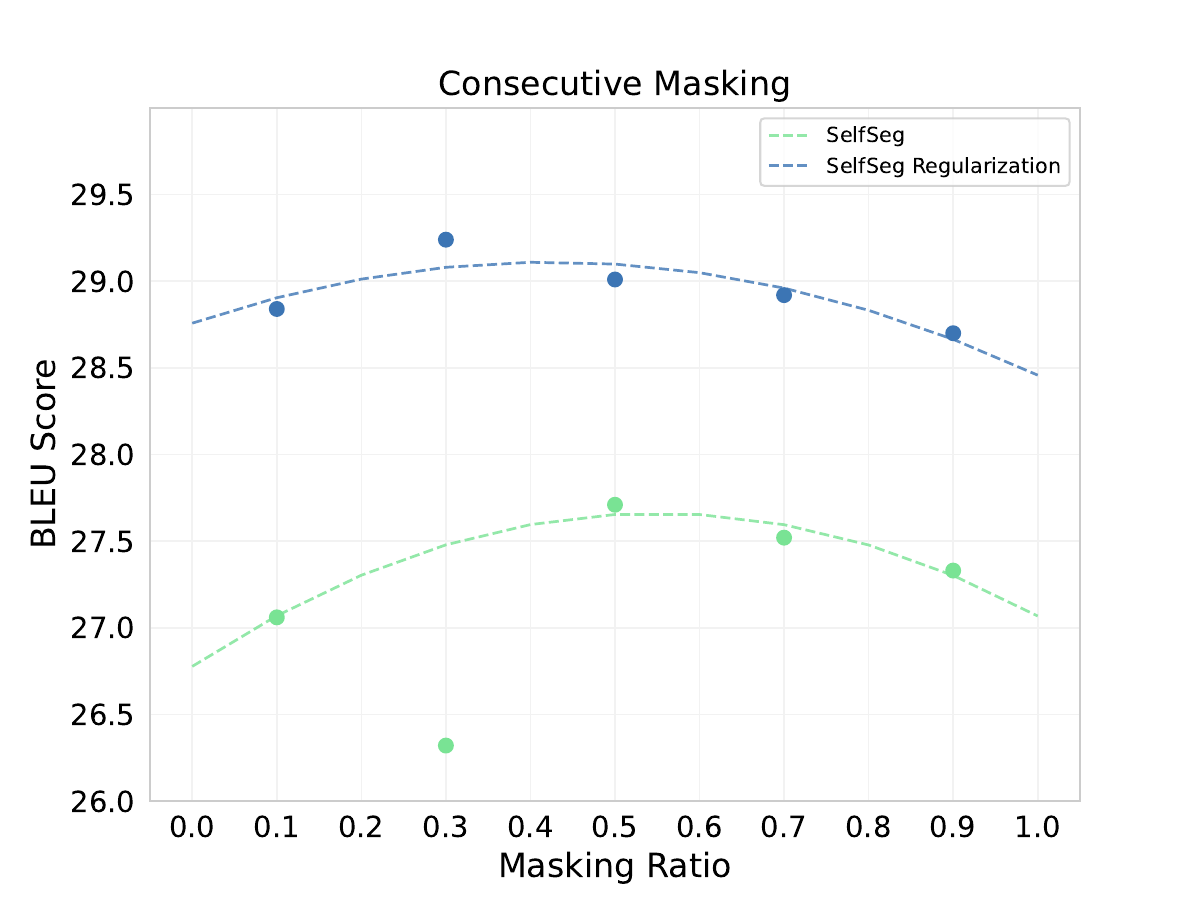}
    \includegraphics[width=0.48\columnwidth]{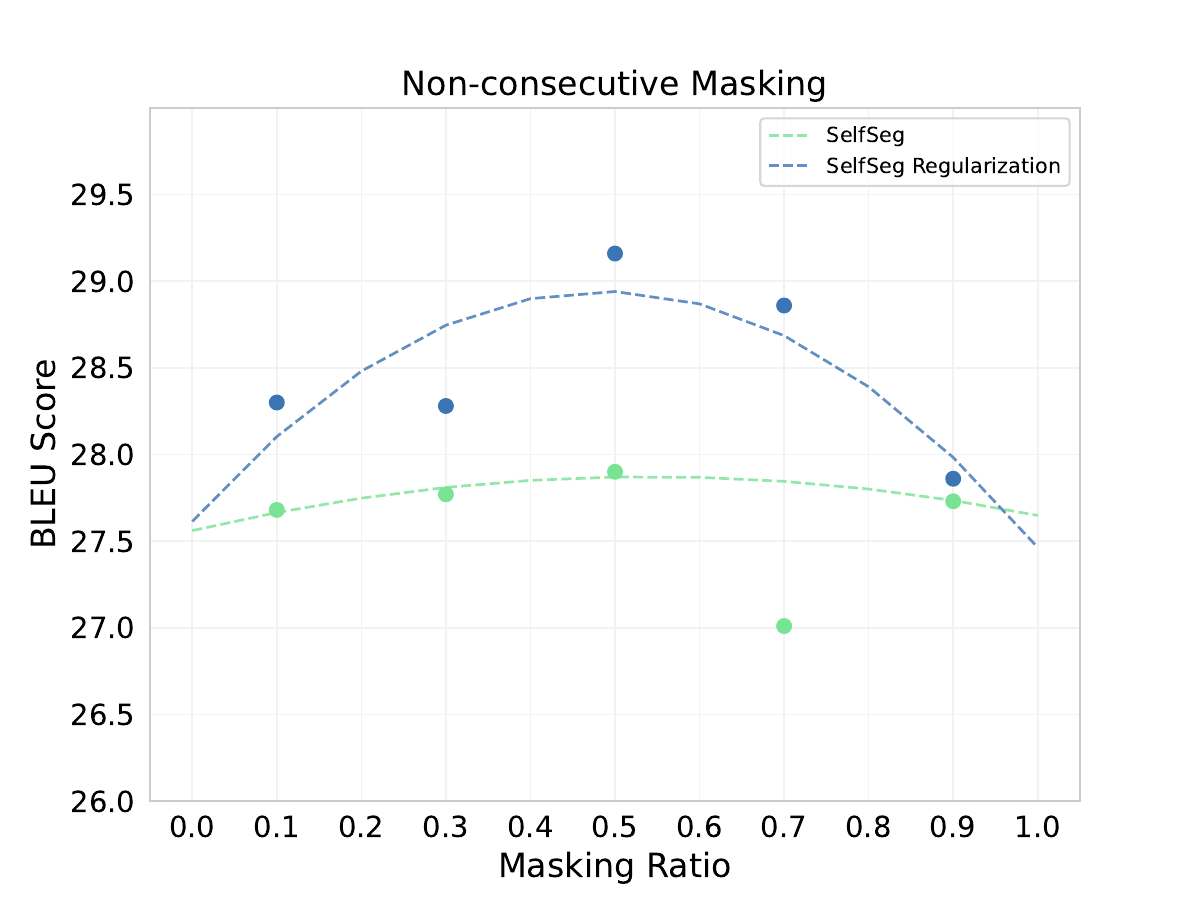}
    \caption{\hl{Performance of using different masking ratios and strategies for charMASS. Left: Consecutive masking strategy. Right: Non-consecutive masking strategy. Tested on the Vi\ra En direction of the IWSLT15 dataset.}}
    \label{fig:maskratio}
\end{figure}

\hl{Furthermore, results of using 1) different masking ratios and 2) consecutive or non-consecutive masking strategies for charMASS on the Vi\ra En direction of the IWSLT15 dataset are shown in \Fig{maskratio}. We mask $\lfloor ratio*\#chars \rfloor$ characters in each word. For the consecutive strategy, we choose the start point of the masking span from the possible start points randomly. For the non-consecutive strategy, we shuffle a list containing $0$s with the number of masking characters and $1$s with the number of non-masking characters to obtain the masking positions.
We found that for both consecutive masking and non-consecutive masking methods, $0.5$ is the best ratio for all settings except SelfSeg-Regularization with consecutive masking, and the performance drops if the masking ratio is very high ($0.9$) or very low ($0.1$). Additionally, there is no significant difference between using consecutive masking and non-consecutive masking strategies.}

\begin{table}[thb]
\small
    \caption{The BLEU scores of SelfSeg with different masking strategies.}
    \begin{tabu}{rSSSSSS|S}
        \toprule
        
         & \textbf{Fil}\ra \textbf{En} & \textbf{Id\ra En} & \textbf{Ja\ra En} & \textbf{Ms\ra En} & \textbf{Vi\ra En} & \textbf{Zh\ra En} & \textbf{Avg} \\ 
        \midrule
        
        \textit{charMASS} &\bfseries 25.20&\bfseries  27.10&\bfseries  11.39&28.15&\bfseries 22.44&\bfseries 12.03&\textbf{21.05} \\
        
        \textit{subwordMASS} & 24.51&26.42&11.17&\bfseries 28.60&21.15&10.97&20.47 \\
        
        \textit{subwordMASK} &23.79&25.35&10.26&28.34&21.37&12.00&20.19 \\
        
        \textit{w/o mask} &24.11&26.45&10.23&28.12&21.20&11.85&20.33\\
        \bottomrule
    \end{tabu}
    \label{tab:masking-strategies}
\end{table}

%

\subsection{Word Frequency Normalization Strategies}
\Tab{word-normalization-strategies} presents the performance of SelfSeg using different word frequency normalization strategies.
We found that 1) using word frequency normalization shows comparable BLEU scores with \textit{w/o Norm}, and 2) all strategies yield similar results except \textit{One}, which may come from the large difference in frequency distribution between training and real data. We used subwordMASS strategy here.

\renewcommand{\arraystretch}{1.3}
\begin{table}[thb]
\small
    \caption{BLEU scores of SelfSeg with different normalization strategies.}
    \begin{tabu}{rSSSSSS|S}
        \toprule
        
        & \textbf{Fil\ra En} & \textbf{Id\ra En} & \textbf{Ja\ra En} & \textbf{Ms\ra En} & \textbf{Vi\ra En} & \textbf{Zh\ra En} & \textbf{Avg} \\ 
        \midrule
        
        \textit{w/o Norm} & 24.51&26.42&11.17&28.60&21.15&10.97&\textbf{20.47}\\
        
        \textit{Threshold} & 24.74&25.86&10.42&28.82&20.90&12.06&\textbf{20.47} \\
        
        \textit{Sqrt}&23.48&25.95&10.64&28.31&20.64&12.04&20.18  \\
        
        \textit{Log} & 24.80&26.03&11.67&28.21&20.67&12.71&20.68 \\
        
        \textit{One} & 22.89&24.80&10.25&26.82&19.87&11.56&19.37 \\
        \bottomrule
    \end{tabu}
    \label{tab:word-normalization-strategies}
\end{table}

\subsection{Types of Training Data}
We demonstrate that parallel or sentence-level training data is unnecessary and monolingual word-level data is sufficient by both sub-word segmentation results and MT results.

\noindent

\textbf{Metrics} The MT performance is measured by BLEU scores, and we measure the difference in sub-word segmentation generated by two segmenters on a given dataset through the following metric.

For each word, we define the Word Difference Rate ($DIF_{word}$) by \Eq{context_agnostic_1}, where $S_1$ and $S_2$ are sets of sub-word segmentations for the given $word$ generated by two segmenters. $|S|$ is the size of $S$, $nword$ is the frequency of $word$ in the corpus.
\begin{align}
\begin{split}
    \mathclap{\mathit{DIF_{word}(S_1, S_2, word)}=\frac{\sum_{i=1}^{|S_1|}\sum_{j=1}^{|S_2|} ({\mathit{seg}_{i}\neq \mathit{seg}_{j}})}{\mathit{nword^2}}}
\end{split}
\label{eq:context_agnostic_1}
\end{align}

We define Corpus Different Rate ($DIF_{corpus}$) based on $DIF_{word}$ in \Eq{context_agnostic_2}, where $W$ is a set containing all types of words $word_i$ for the given corpus, $|W|$ is the size of $W$.
\begin{align}
\begin{split}
     \mathit{DIF_{corpus}(S_1, S_2, W)}=\frac{\sum_{i=1}^{|W|} \mathit{DIF_{word}(S_1, S_2, word_i)}*\mathit{nword}_i}{\sum_{i=1}^{|W|} \mathit{nword}_i}
\end{split}
\label{eq:context_agnostic_2}
\end{align}

Additionally, if $S_1=S_2$, $DIF_{word}$ and $DIF_{corpus}$ measure the consistency of segmentation of the same word in different sentences by the segmenter.

\noindent

\textbf{Settings} We calculate $DIF_{corpus}$ among BPE, DPE, SelfSeg-Sentence (using sentence-level data), and SelfSeg on the English part of the IWSLT'15 Vi-En dataset. All four methods use the same vocabulary. The input of SelfSeg-Sentence is a monolingual sentence instead of word during both training and decoding. subwordMASS is used for SelfSeg and SelfSeg-Sentence.


\noindent

\begin{figure}[htb]
    \centering  
    \includegraphics[width=0.55\columnwidth]{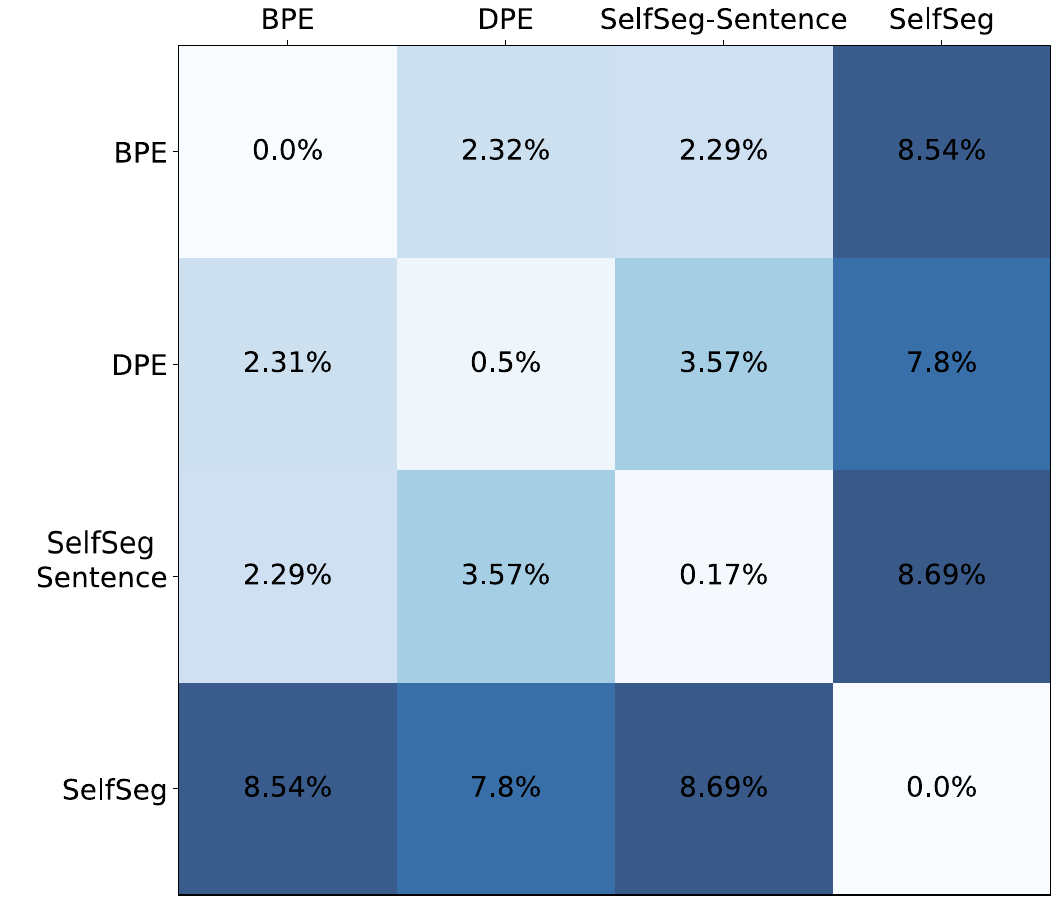}
    \caption{$DIF_{corpus}$ among different segmentation methods on the IWSLT'15 Vi-En dataset.}
    \label{fig:similarity}
\end{figure}

\textbf{Parallel Data is Not Necessary}
Tables~\ref{tab:sentence-level-res} and~\ref{tab:sentence-level-res-iwslt} present the MT results where using monolingual sentence-level data achieved higher BLEU scores than using parallel data. 
\Fig{similarity} shows the $DIF_{corpus}$ results. SelfSeg-Sentence gives more consistent segmentations compared with DPE ($0.17\%$ vs. $0.5\%$).

\noindent

\textbf{Sentence-level Data is Not Necessary}
Comparing SelfSeg-Sentence (DPE) and SelfSeg, we can find that SelfSeg using word-level data achieves higher MT performance, showing that word-level data is enough for MT. The DPE work~\citep{he-etal-2020-dynamic} used sentence-level data based on the assumption that one word will be segmented differently in different contexts. However, we found the $DIF_{corpus}(DPE, DPE, IWSLT15)$ is only $0.5\%$ percentage, showing that this assumption is not valid. Furthermore, we divided the words occurring in the dataset into two sets, $W_{high}$ containing high-frequency words ($nword>5$) and $W_{low}$ containing low-frequency words ($nword<=5$). We found $DIF_{corpus}(DPE, DPE, W_{high})$ is only $0.40\%$ whereas $DIF_{corpus}(DPE, DPE, W_{low})$ is $6.14\%$. Even for the $W_{low}$ with high $DIF_{corpus}$, one word should be segmented consistently.
For example, DPE segments word \textit{jumbled} into \textit{ju+mble+d} and \textit{j+umb+led}, word \textit{mended} into \textit{me+nded} and \textit{m+ended}, whereas the SelfSeg generates \textit{j+umb+l+ed} and \textit{m+end+ed}.

\begin{table}[thb]
    \caption{BLEU scores on the ALT dataset using segmenters trained on different types of data. DPE uses parallel sentence-level data, SelfSeg-Sentence uses monolingual sentence-level data, and SelfSeg uses monolingual word-level data.}
    \resizebox{0.78\textwidth}{!}{
    \begin{tabu}{cSSSSSS|S}
        \toprule
        
        & \textbf{Fil\ra En} & \textbf{Id\ra En} & \textbf{Ja\ra En} & \textbf{Ms\ra En} & \textbf{Vi\ra En} & \textbf{Zh\ra En} & \textbf{Avg} \\ 
        \midrule
        
       BPE~\cite{sennrich-etal-2016-neural} & 23.09&25.70&9.42&28.19&19.94&12.21&19.76\\
        
        DPE~\citep{he-etal-2020-dynamic}&24.04&26.66&9.93&27.89&20.06&10.72&19.88\\
        
        SelfSeg-Sentence & 24.28&25.37&10.74&28.25&21.36&12.11&20.35\\
        
        SelfSeg&24.51&26.42&11.17&28.60&21.15&10.97&\bfseries 20.47\\
        \bottomrule
    \end{tabu}
    }
    \label{tab:sentence-level-res}
\end{table}

\renewcommand{\arraystretch}{1.3}
\begin{table}[thb]
\small
    \caption{BLEU scores on IWSLT15 Vi\ra En dataset with segmenters trained on different types of data.
    }
    \begin{tabu}{cS}
        \toprule
        
        & \textbf{IWSLT15 Vi\ra En} \\
        \midrule
        
        BPE~\cite{sennrich-etal-2016-neural} & 27.09 \\
        
        DPE~\citep{he-etal-2020-dynamic}&27.40\\
        
        SelfSeg-Sentence & 27.79\\
        
        SelfSeg&\bfseries 28.19\\
        \bottomrule
    \end{tabu}
    \label{tab:sentence-level-res-iwslt}
\end{table}

\subsection{Sizes of Training Data for SelfSeg}
\hl{In this section, we investigate the impact of the amount of monolingual data used in the segmenter training. The results are represented in \Tab{dataset-sizes}. The amount of English data to train the SelfSeg segmenter varies from $18k$ to $10M$, where the $18k$ setting used the ALT English data, the $50k$ and $532k$ setting used the news commentary corpus, the $4.5M$ setting used the English side of the WMT14 De-En dataset and the $10.0M$ setting used the English side of the WMT14 Fr-En dataset. We find that using more monolingual data brings performance improvement. Especially with $10$M monolingual sentences from WMT14 Fr-En, the improvement reached $1.7$ BLEU score compared with SelfSeg using $18k$ monolingual sentences. Although with more data, the performance of using BPE also improves, the improvement is small compared with that of SelfSeg.}


\renewcommand{\arraystretch}{1.3}
\begin{table}[thb]
    \caption{Performance of the segmenter model trained on different sizes of the training data. We only have DPE 18k because it uses ALT parallel data.
    }
    \resizebox{0.78\textwidth}{!}{
    \begin{tabu}{cSSSSSS|S}
        \toprule
        
        & \textbf{Fil\ra En} & \textbf{Id\ra En} & \textbf{Ja\ra En} & \textbf{Ms\ra En} & \textbf{Vi\ra En} & \textbf{Zh\ra En} & \textbf{Avg} \\ 
        \midrule
        
        \multicolumn{8}{l}{\textit{Size: 18k}} \\
        
        BPE~\cite{sennrich-etal-2016-neural} & 23.09&25.70&9.42&28.19&19.94&12.21&19.76 \\
        
        DPE~\citep{he-etal-2020-dynamic}&24.04&26.66&9.93&27.89&20.06&10.72&19.88\\
        
        SelfSeg&24.11&25.85&11.11&28.73&20.68&10.46&\bfseries 20.16\\
        \midrule
        
        \multicolumn{8}{l}{\textit{Size: 50k}} \\
        
        BPE~\cite{sennrich-etal-2016-neural}&23.90&25.62&10.54&28.62&19.99&11.51&20.03\\
        
        SelfSeg&24.51&26.42&11.17&28.60&21.15&10.97&\bfseries 20.47\\
        \midrule
        \rowfont{\color{blue}}
        \multicolumn{8}{l}{\textit{Size: 532k}} \\
        \rowfont{\color{blue}}
        BPE~\cite{sennrich-etal-2016-neural}& 23.71&25.68&10.59&28.53&21.57&11.04&\textbf{20.19}\\
        \rowfont{\color{blue}}
        SelfSeg&23.96&26.40&9.93&28.01&20.92&11.66&20.15\\
        \midrule
        \rowfont{\color{blue}}
        \multicolumn{8}{l}{\textit{Size: 4.5M}} \\
        \rowfont{\color{blue}}
        BPE~\cite{sennrich-etal-2016-neural}& 23.77&26.43&10.64&27.75&22.11&11.37&20.35\\
        \rowfont{\color{blue}}
        SelfSeg&25.60&26.45&10.09&28.57&20.23&12.22&\textbf{20.53}\\
        \midrule
        \rowfont{\color{blue}}
        \multicolumn{8}{l}{\textit{Size: 10.0M}} \\
        \rowfont{\color{blue}}
        BPE~\cite{sennrich-etal-2016-neural}&24.53&25.69&10.59&28.14&21.45&11.92&20.39\\
        \rowfont{\color{blue}}
        SelfSeg&26.49&27.37&11.84&29.47&23.10&12.90&\textbf{21.86}\\
        \bottomrule
    \end{tabu}
    }
    \label{tab:dataset-sizes}
\end{table}

\subsection{Lightweight SelfSeg Model}
We examine a lightweight segmenter model SelfSeg-Light, given that the training data is word-level and on a small scale.
The architecture of SelfSeg-Light is a single-layer transformer encoder and a single-layer transformer decoder.
As illustrated in \Tab{light-model}, the performance of SelfSeg-Light is comparable with SelfSeg, which indicates that maybe there is no need to use a large model.

\renewcommand{\arraystretch}{1.3}
\begin{table}[thb]
\small
    \caption{BLEU scores of transformers with 4- and 1-layer. With charMASS strategy.}
    \begin{tabu}{cSSSSSS|S}
        \toprule
        
        & \textbf{Fil\ra En} & \textbf{Id\ra En} & \textbf{Ja\ra En} & \textbf{Ms\ra En} & \textbf{Vi\ra En} & \textbf{Zh\ra En} & \textbf{Avg} \\ 
        \midrule
        
        SelfSeg&25.20&27.10&11.39&28.15&22.44&12.03& \bfseries 21.05 \\
        
        SelfSeg-Light&25.26&26.07&11.34&29.10&20.64&12.05&20.74\\
        \bottomrule
    \end{tabu}
    \label{tab:light-model}
\end{table}

\subsection{Segmentation Case Study}
\label{segmentation_case_study}
In this section, we analyze the segmentation and show why the segmentation generated by our method leads to better performance on the downstream MT task.

Table~\ref{tab:seg_examples} shows examples of words with different segmentations between the BPE and SelfSeg method on the ALT dataset. We can observe that the BPE method tends to generate high-frequency sub-words, due to the greedy strategy, whereas our SelfSeg, powered by the DP algorithm, tends to generate linguistically intuitive combinations of sub-words for not only frequent words but also rare words.
This observation is similar to that by \cite{he-etal-2020-dynamic}. Additionally, \Tab{seg_regularization_examples} provides some examples of sub-word segmentations by BPE-dropout~\cite{provilkov-etal-2020-bpe} and proposed SelfSeg-Regularization. Both methods yield high diversity of segmentations while the proposed method generates more linguistically intuitive sub-words.

To verify whether our segmentation looks intuitive for the neural models, we trained neural word language models with $Transformer_{base}$ architecture,\fn{\url{https://github.com/pytorch/examples/tree/master/word\_language\_model}} the same used in the MT experiments, and checked the decoding perplexity. 
For each segmentation method, we train an English neural word language model on the ALT-train set and test on the ALT-test set segmented by that method. As presented in \Tab{seg_LM_res}, the decoding perplexity of DPE and SelfSeg methods are much lower than that of the BPE method, which we assume is due to the optimization of the log marginal
likelihood of the DPE method. From the results of neural LMs, we can infer that when applying our segmentations to MT tasks, the decoder tends to be more certain, as indicated by the low entropy.

\begin{table}[!thb]
\small
    \caption{\textbf{Examples of sub-word segmentations by different approaches.} The frequency of rare words are $<5$, and one-shot words appear only once in the ALT training data.}
    \begin{center}
    \begin{tabular}{ll|ll}
        \toprule
        \textbf{BPE}~\cite{sennrich-etal-2016-neural} & \textbf{SelfSeg} & \textbf{BPE}~\cite{sennrich-etal-2016-neural} & \textbf{SelfSeg} \\
        \midrule
        \multicolumn{2}{c}{\textit{frequent words}} & \multicolumn{2}{c}{\textit{rare words}} \\
        dam + aged&damage + d & d + raf + ting & d + raft + ing \\
        com + ments&comment + s & murd + ered & murder + ed\\
        hous + es&house + s & Net + w + orks & Net + work + s\\
        subsequ + ently & subsequent + ly & aut + h + ored & author + ed\\
        wat + ching&watch + ing & disag + reed & disagree + d\\
        sec + retary&secret + ary & \multicolumn{2}{c}{\textit{one-shot words}} \\
        un + k + n + own&un + know + n & reinfor + ces & reinforce + s \\
        refere + es&refer + ee + s & sub + stit + utions & sub + stitution + s\\
        langu + ages&language + s  & trad + em + ar + ks & trade + mark + s\\
        you + n + gest & young + est & ris + king & risk + ing\\
        mom + ents & moment + s & Somet + hing & Some + thing\\
        \bottomrule
    \end{tabular}
    \label{tab:seg_examples}
    \end{center}
\end{table}

\begin{table}[!thb]
\small
    \caption{Examples of sub-word segmentations by BPE-dropout and SelfSeg-Regularization on the ALT dataset.}
    \begin{center}
    \begin{tabular}{llll}
        \toprule
        \textbf{BPE-dropout}~\cite{provilkov-etal-2020-bpe} & \textbf{SelfSeg-Regularization}\\
        \midrule
        \multicolumn{2}{c}{\textit{frequent words}}\\
        sub + sequ + ently & subsequent + ly\\
        subsequ + ently & subsequ + ent + ly\\
        s + ub + sequ + ent + l + y & subsequent + l + y\\ 
        sub + sequ + ently & subsequent + ly\\
        subsequently & subsequ + en + t + ly\\
        \multicolumn{2}{c}{\textit{rare words}}\\
        disag + reed & disagree + d \\
        d + is + ag + reed & disag + r + e + ed \\
        disag + re + ed & disag + re + e +d \\
        disag + reed &  dis + ag + r + e + ed \\
        d + is + ag + reed & disagree + d \\
        \multicolumn{2}{c}{\textit{one-shot words}}\\
        rein + for + ces & reinforce + s \\
        re + in + f + or + ces & reinfor + ces \\
        re + in + for + ces &  reinforce + s \\
        rein + for + ces & reinforce + s \\
        re + in + for + ces & reinfor + c + e + s\\
        \bottomrule
    \end{tabular}
    \label{tab:seg_regularization_examples}
    \end{center}
\end{table}

\begin{table}[thb]
\small
    \caption{The perplexity of neural LMs. SelfSeg uses subwordMASS strategy.}
    \begin{center}
    \begin{tabu}{crrr}
        \toprule
        &  \textbf{PPL per line} & \textbf{PPL per token} & \textbf{\# tokens per line} \\
        \midrule
        BPE~\cite{sennrich-etal-2016-neural}& 29,688.2 & 799.4 & 37.1 \\
        DPE~\citep{he-etal-2020-dynamic} & 28,498.7 & 816.1 & 34.9 \\
        SelfSeg & 28,714.0 & 772.2 & 37.2 \\
        \bottomrule
    \end{tabu}
    \label{tab:seg_LM_res}
    \end{center}
\end{table}

\section{Conclusion and Future Work}
We proposed a novel method SelfSeg for neural sub-word segmentation to improve the performance of NMT and only requires monolingual word-level data. 
It models the word generation probability through all segmentations and chooses the segmentation with MAP. 
We propose masking strategies to train the model in a self-supervised manner, word-frequency normalization methods to improve the training speed, and a regularization mechanism that helps to generate segmentations with more variety.
Experimental results show that NMT using proposed SelfSeg methods is either comparable to or better than NMT using BPE and DPE in low-resource to high-resource settings And the regularization mechanism achieves a large improvement over baseline methods.

Furthermore, both the training speed and testing speed are more than ten times faster than those of DPE. 
Analyses show the context agnostic property of the sub-word segmentation, therefore sentence-level training data is not required. 
Moreover, the segmentations given by the proposed method are more linguistically intuitive as well as easier for the neural decoder to generate as indicated by the low entropy. 

Our future work will focus on several directions. 
First, we are implementing the pre-trained encoder such as BERT/mBERT/BART/mBART on the segmenter. The charMASS method only captures the lexical information and involving semantic information may further improve the quality.
Second, we will try to extend the model to multilingual settings. In this way, we only need to train one model to pre-process data of all languages instead of training multiple models for different languages, which can drastically reduce the training time and increases the efficiency of the application. 
Third, the direction of joint training of the segmenter and the downstream tasks model is also promising, where the segmenter will be aware of the downstream tasks explicitly and be optimized to improve the performance of downstream tasks.
Finally, optimizing the vocabulary for sequence generation is necessary. Although the segmentations are optimized for the neural model to generate the word, the possible segments themselves are generated by BPE, which are not optimized for sequence generation. 

\begin{acks}
This work was done during the internship in National Institute of Information and Communications Technology.
This work was supported by JSPS KAKENHI Grant Number $21J23124$.
This work was also supported by a Grant-in-Aid for Young Scientists \#19K20343, JSPS, and JSPS Research Fellow for Young Scientists (DC1).
\end{acks}

\bibliographystyle{ACM-Reference-Format}
\bibliography{reference}

\appendix

\end{document}